%% file: ACMKDD_SocraSynth.tex
\documentclass[sigconf,nonacm]{acmart}

\usepackage{tikz}
\usepackage{multicol}
\usepackage{graphicx}
\usepackage{multirow}
\usepackage{enumitem}
\usepackage{tabularx}
\usepackage{fancyvrb} 
\usepackage{balance} 
\title{SocraSynth: Multi-LLM Reasoning with Conditional Statistics}

\author{Edward Y. Chang}
\affiliation{
  \institution{Computer Science, Stanford University}
  \city{}
  \country{}}
\email{echang@cs.stanford.edu}

\begin{abstract}
Large language models (LLMs), while promising, face criticisms for biases, hallucinations, and a lack of reasoning capability. This paper introduces SocraSynth, a multi-LLM agent reasoning platform developed to mitigate these issues. SocraSynth utilizes conditional statistics and systematic context enhancement through continuous arguments, alongside adjustable debate contentiousness levels. The platform typically involves a human moderator and two LLM agents representing opposing viewpoints on a given subject. SocraSynth operates in two main phases: knowledge generation and reasoning evaluation. In the knowledge generation phase, the moderator defines the debate topic and contentiousness level, prompting the agents to formulate supporting arguments for their respective stances. The reasoning evaluation phase then employs Socratic reasoning and formal logic principles to appraise the quality of the arguments presented. The dialogue concludes with the moderator adjusting the contentiousness from confrontational to collaborative, gathering final, conciliatory remarks to aid in human reasoning and decision-making. Through case studies in three distinct application domains, this paper showcases SocraSynth's effectiveness in fostering rigorous research, dynamic reasoning, comprehensive assessment, and enhanced collaboration. This underscores the value of multi-agent interactions in leveraging LLMs for advanced knowledge extraction and decision-making support.
\end{abstract}

\keywords{Collaborative reasoning, conditional statistics, large language models, Socratic method, SocraSynth}

\newcommand{\BibTeX}{\rm B\kern-.05em{\sc i\kern-.025em b}\kern-.08em\TeX}

\acmConference[KDD'24]{ACM KDD Conference}{Barcelona}{Spain}
\setcopyright{none}
\renewcommand\footnotetextcopyrightpermission[1]{}

\setcopyright{none}
\makeatletter
\renewcommand\@formatdoi[1]{\ignorespaces}
\makeatother

\begin{document}

\pagestyle{plain}

\maketitle 

\input{Introduction.tex}

\input{Overview}

\input{CaseStudyPolicy}
\input{CaseStudyHealth}
\input{CaseStudyContentiousness}

\input{Conclusions}

\bibliographystyle{ACM-Reference-Format} 
\bibliography{References, SocraHealth}
\input{AppendixLong}
\input{AppendixHealth}

\end{document}

%% file: Introduction.tex
\section{Introduction}
\label{sec:intro}

Revolutionary advancements in large language models (LLMs) \cite{Bbubeck2023sparks,thoppilan2022lamda,touvron2023llama,chatgpt,geminiteam2023gemini} and, more broadly, foundation models (FMs) \cite{bommasani2022opportunities} have paved the way for multi-agent systems to achieve remarkable progress in knowledge acquisition and natural language understanding \cite{zhang2023cumulative}. As highlighted by \cite{ChangYEdward072023,openai2023gpt4,Bbubeck2023sparks}, 
models such as GPT-4 exhibit information processing
qualities surpassing human capabilities, including:
1) deep and extensive knowledge, 2)
interdisciplinary assimilation and fusion of knowledge, and 3)
multimodal and multilingual expertise.

While promising, LLMs face criticisms
for biases, hallucinations, and 
a lack of reasoning capability \cite{huang2023Hallucinationsurvey}.
To mitigate this disparity, we introduce SocraSynth, a pioneering platform 
blending the principles of ``Socratic Synthesis'' and ``Socratic Symposium.'' 
It fosters collaboration between humans and LLM agents, enabling the formation of deep questions and transcending typical human constraints in reasoning, validation, and assessment. 
A typical SocraSynth ensemble comprises a human moderator paired with two LLM agents, 
each espousing divergent perspectives. For instance, one side may advocate for regulating AI, 
while the other may oppose such regulation. An agent can be powered by an LLM such as 
GPT-4 \cite{Bbubeck2023sparks}, Gemini \cite{geminiteam2023gemini}, or Llama \cite{touvron2023llama}. 
While the human moderator sets the thematic boundaries of a debate, they refrain from directly contributing to content generation, maintaining impartiality in the deliberations. The tunable 'contentiousness' parameter plays a key role in modulating the debate dynamics, allowing for a range of interactions from adversarial to collaborative.


The operation of SocraSynth unfolds in two main stages: the generative and the evaluative. 
The generative phase sees LLM agents crafting and countering arguments, iterating under the moderator-defined subject until a holistic conclusion is drawn. The evaluative phase, conversely, harnesses diverse virtual judges—each backed by a unique LLM---to impartially assess the debate. The Critical Inquisitive Template (CRIT) algorithm \cite{SocreticMethodChang2023}, rooted in Socratic reasoning \cite{SocraticMethidWiki,PaltoRepublicURL,Paul2007CriticalTT,Irony2022}, serves as the evaluative anchor.

Following these stages, SocraSynth fine-tunes the ``contentiousness'' parameter to encourage a balanced proposal from LLM agents. This proposal, readied for human assessment, encapsulates the combination of multi-agent knowledge discovery and intricate reasoning. This becomes especially prominent in domains centered on open-ended decision-making. Here, the principle of ``reasonableness'' often supersedes clear-cut ``truths,'' especially when such ``truths''---like the question of ``Should Musk have bought Twitter?''---are subjective or intangible.

SocraSynth addresses the challenges of biases, hallucinations, and a lack of reasoning in LLMs through a structured LLM-dialogue framework characterized by conditional statistics, contentiousness modulation, context refinement, and reasonableness evaluation. Conditional statistics are realized by having two LLM agents argue from opposing viewpoints on a topic. This approach inherently challenges the default biases of the LLMs, as each agent must consider and defend a stance that may contrast with the biases present in their training data.

Contentiousness modulation plays a crucial role in this process. The debate begins with the agents adopting polarized positions on the subject matter. As the dialogue progresses, the tone shifts from
confrontational to more collaborative, culminating in conciliatory remarks. This progression helps unearth diverse and nuanced perspectives, countering the potential biases of a single-agent, monologue-based question-answering system.

To combat the issue of hallucinations---instances where LLMs generate irrelevant or nonsensical content---SocraSynth utilizes iterative dialogue rounds. Each round, composed of arguments and counterarguments, continuously refines the context of the debate. This dynamic, back-and-forth interaction significantly reduces the likelihood of irrelevant or illogical responses, as each agent's input is constantly checked and challenged by the other.

Lastly, the evaluation of ``reasonableness'' is a
critical step post-debate. SocraSynth employs the Socratic method to assess each argument's logical validity and source credibility. This analysis is not just a surface-level check; it delves into the foundations of each argument, scrutinizing its rationality and the reliability of its sources. The human mediator, or moderator, then uses this evaluation to provide targeted feedback to both LLM agents. This feedback loop is essential for refining the agents' reasoning capabilities and enhancing their ability to generate credible, logically sound arguments.

SocraSynth has demonstrated successful applications across diverse fields, such as geopolitical analysis \cite{MiddleEast-October2023}, medical diagnostics \cite{SocraHealth2023}, sales strategy formulation \cite{SocraPlan2023}, and Wikipedia article enhancement \cite{SocraPediaDecember2023}. These applications have notably expanded perspectives and enhanced argumentation quality. There is also substantial evidence of a significant reduction in biases and hallucinations within these varied domains, highlighting SocraSynth's effectiveness in creating balanced and well-reasoned discourse.

The remainder of this paper delves into SocraSynth's architectural intricacies, algorithmic underpinnings, and real-world impacts in greater detail. The key contributions of this paper include:

\begin{enumerate}[left=0em,label=\arabic*.]
\item Introduction of the SocraSynth framework, designed to amplify interdisciplinary reasoning with LLMs, and featuring unique algorithmic elements such as conditional statistics for generating balanced arguments.
\item An in-depth exploration of the contentiousness modulation algorithm, a crucial SocraSynth feature that dynamically adjusts debate intensity, enabling a range from confrontational to collaborative interactions.
\item Implementation of context refinement within SocraSynth, which continuously enhances the relevance and accuracy of arguments produced by LLM agents, thereby elevating the overall quality of the discourse.
\item Development and incorporation of the reasonableness evaluation mechanism, essential for assessing the logical soundness and source credibility of arguments, thus ensuring the integrity and utility of the discussions.
\end{enumerate}

Collectively, these four algorithmic innovations—conditional statistics, contentiousness modulation, context refinement, and reasonableness evaluation—constitute the core of the SocraSynth framework. This integration not only represents technical ingenuity but also significantly contributes to the broader landscape of AI-driven debates. By effectively addressing challenges like biases and hallucinations, SocraSynth sets a new benchmark in knowledge discovery and reasoning, underscoring its profound impact on the field of artificial intelligence and beyond.


%% file: Overview.tex
\section{Multi-Agent SocraSynth Overview}
\label{sec:Overview}

\begin{figure}[t!]
  \centering
  \includegraphics[width=0.88\linewidth]{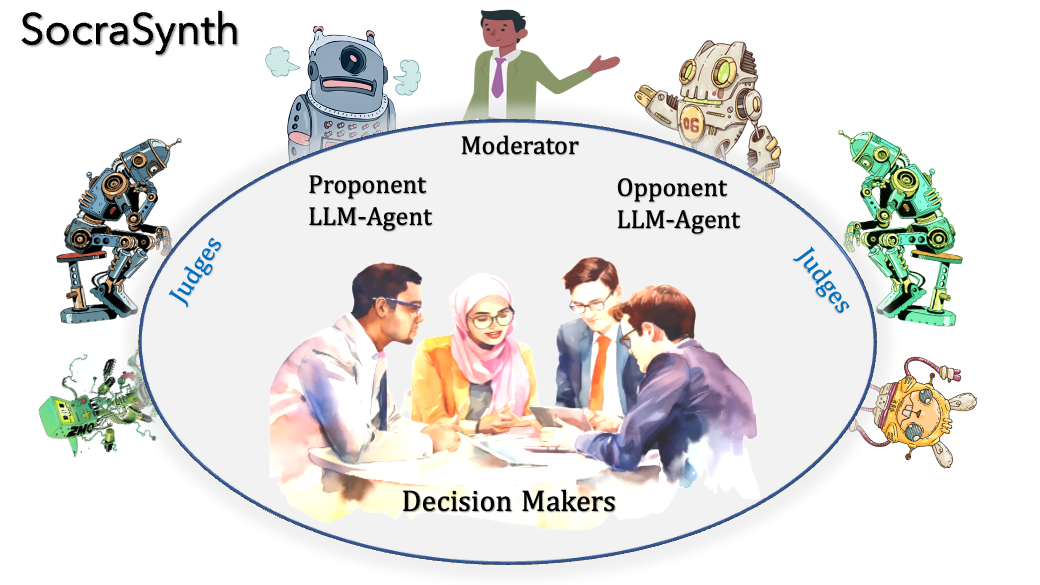}
  \vspace{-.1in}
  \caption{SocraSynth Agents and Roles.}
  \label{fig:socrasynth}
  \Description{SocraSynth Agents and Roles.}
  \vspace{-.3in}
\end{figure}

SocraSynth is a multi-agent collaborative reasoning platform that skillfully integrates human intelligence with the capabilities of Large Language Model (LLM)-powered agents. As illustrated in Figure~\ref{fig:socrasynth}, each participant plays a vital role: humans act as moderators, LLM agents are responsible for generating knowledge, LLM judges conduct evaluations, and human executives make the final decisions. The integration of LLMs significantly boosts the platform's effectiveness, leveraging their extensive knowledge bases and extraordinary interdisciplinary reasoning abilities. An LLM can be thought of as an entity possessing expertise across a multitude of fields, akin to holding Ph.D.s in various disciplines, enabling it to seamlessly navigate and synthesize a wide range of knowledge.

Engaging with an LLM is comparable to a scenario where a 10-year-old joins a scholarly discussion with a group of Nobel Laureates. The disparity in knowledge and experience is considerable, posing a significant challenge for the younger participant to engage meaningfully in such advanced intellectual discourse. In this analogy, expecting the 10-year-old, or anyone with limited expertise, to pose profound questions that elicit insightful answers is unrealistic. SocraSynth addresses this disparity by shifting the paradigm: instead of having the less informed individuals pose questions, it orchestrates a debate among the Nobel Laureates, or LLMs, with humans assuming the role of moderators.

This approach not only addresses the challenge of asymmetric knowledge but also resolves critical issues such as model biases and hallucination challenges inherent in LLMs. Within SocraSynth, a human moderator initiates the topic for discussion or debate. LLM agents, each embodying different perspectives, contribute their knowledge, potentially revealing new insights that the moderator might be unaware of. This diverse representation helps counteract the model biases that often arise from training data, as each LLM agent is encouraged to explore and present varying viewpoints. During and after 
the debate, another set of diverse LLM agents undertakes impartial evaluations. This step is crucial 
in mitigating hallucinations ---instances where LLMs generate
irrelevant or nonsensical content. By incorporating a variety 
of agents for evaluation, SocraSynth ensures that the 
content produced during the debate is critically examined 
for its relevance and coherence, further reducing the likelihood of hallucinatory responses. 

\begin{table*}[t!]
\vspace{-.1in}
\centering
\begin{small}
\begin{tabular}{|p{0.48cm}|p{4.6cm}|p{4.6cm}|p{4.6cm}|}\toprule
\textbf{C.L.} & \textbf{Tone} & \textbf{Emphasis} & \textbf{Language} \\
\hline \hline
$0.9$ & Highly confrontational; focused on raising strong ethical, scientific, and social objections. & Highlighting risks and downsides; ethical quandaries, unintended consequences, and exacerbation of inequalities. & Definitive and polarizing, e.g., ``should NOT be allowed,'' ``unacceptable risks,'' ``inevitable disparities.'' \\
\hline
$0.7$ & Still confrontational but more open to potential benefits, albeit overshadowed by negatives. & Acknowledging that some frameworks could make it safer or more equitable, while cautioning against its use. & Less polarizing; ``serious concerns remain,'' ``needs more scrutiny.'' \\
\hline
0.5 & Balanced; neither advocating strongly for nor against gene editing. & Equal weight on pros and cons; looking for a middle ground. & Neutral; ``should be carefully considered,'' ``both benefits and risks.'' \\
\hline
0.3 & More agreeable than confrontational, but maintaining reservations. & Supportive but cautious; focus on ensuring ethical and equitable use. & Positive but careful; ``transformative potential,'' ``impetus to ensure.'' \\
\hline
0.0 & Completely agreeable and supportive. & Fully focused on immense potential benefits; advocating for proactive adoption. & Very positive; ``groundbreaking advance,'' ``new era of possibilities.'' \\
\bottomrule
\end{tabular}
\end{small}
\vspace{.03in}
\caption{Changes in Arguments at Different Contentiousness Levels.}
\vspace{-.3in}
\label{tab-contentiousness}
\end{table*}

The operational framework of SocraSynth, thus, is bifurcated into two main stages: 
the {\em generative} stage, where knowledge is created and exchanged in a debated format, 
and the {\em evaluative} stage, which focuses on assessing 
the quality and validity of the arguments presented. 
This dual-stage structure, elaborated upon in subsequent sections, 
is instrumental in overcoming the limitations of LLMs by providing
a comprehensive platform for not only generating diverse viewpoints but also critically examining and refining these viewpoints to ensure their logical soundness and relevance. Through this innovative design, SocraSynth effectively navigates the challenges posed by model biases and hallucinations, enhancing the reliability and depth of knowledge extraction and reasoning processes.

\subsection{Generative Stage}
\label{sec:SocraSynth-Generative}

In the generative stage of SocraSynth, LLM agents partake in intensive debates, delving into the nuances of the given topic. This vibrant interaction plays a key role in fostering thorough intellectual discourse, bringing to light the complexities of the subject matter. The CRIT algorithm, which will be detailed in Section~\ref{sec:SocraSynth-Evaluative}, is employed to evaluate the quality of these arguments.

While the generative phase of SocraSynth does not adhere to strict logical frameworks such as first-order logic, it excels in distributed reasoning. This process involves a progressive exchange of arguments and counterarguments, allowing for the gradual honing and refinement of ideas. Open-domain logical reasoning, as described by \cite{bommasani2022opportunities}, demands logical deductions from a wide range of data sources. SocraSynth, leveraging the comprehensive capabilities of GPT-4 as demonstrated in the MMLU benchmark \cite{hendrycks2021measuring,Bbubeck2023sparks}, integrates various NLP functions to facilitate this reasoning process.

In this context, the series of arguments and counterarguments effectively function as targeted questions and answers, each with a clear goal, question, and contextual framework. Through iterative dialogue rounds on each sub-topic, SocraSynth significantly reduces the chances of misunderstanding questions and contextual information, ensuring clarity and precision in the discourse.

\subsubsection{\bf Mitigating Model Biases}

In shaping the nature of debate within SocraSynth, the contentiousness parameter is instrumental. It compels LLM agents to consider and represent a range of perspectives, particularly those that are typically underrepresented or more polarized with respect to the discussion topic. This strategic approach mitigates the inherent biases that arise from the training data of LLMs and guides the discourse towards a wider and more varied exploration of ideas.

Table~\ref{tab-contentiousness} demonstrates how altering the contentiousness levels results in marked changes in GPT-4's tone and approach. (The details
of the experiment is presented in Section~\ref{sec-experiments}.) A high contentiousness level, such as 0.9, leads to highly confrontational interactions, with GPT-4 presenting strong objections and emphasizing the negatives through polarizing language. Conversely, as the contentiousness is reduced, GPT-4's tone shifts to a more conciliatory demeanor, acknowledging potential benefits and considering alternative perspectives, thus fostering a more cooperative dialogue.

The modulation of the contentiousness parameter within 
the generative stage is a crucial mechanism for SocraSynth to mitigate model biases inherent in LLMs due to their training data. By adjusting levels of contentiousness, SocraSynth compels LLMs to venture beyond 
their {\em default} positions---much like a vegetarian, when faced with no other choice, might be compelled to consume meat. In this way, LLMs are {\em freed} from their typical statistical leanings, enabling them to articulate a spectrum of arguments that spans from highly contentious to conciliatory. This not only diversifies the discourse but also ensures that the debate encompasses a full range of perspectives. Consequently, this process allows LLMs to generate responses that break free from the constraints of their training, fostering the emergence of novel and less predictable ideas in the conversation.

\subsubsection{\bf Eliminating Hallucination}

Further, the iterative nature of the debates within SocraSynth cultivates a ``reasonableness'' in information discovery that conventional logical methods may not achieve. Through persistent reasoning and the critical assessment of claims, LLM agents refine their arguments iteratively. This structured debate format significantly diminishes the chance of erroneous claims persisting. Considering that the likelihood of two agents aligning on a false premise is extremely low, the SocraSynth debate format effectively ensures the intellectual integrity of the discourse and substantially reduces the risk of perpetuating fallacies or hallucinations. This methodical refinement process, facilitated by continuous argumentation and opposition, underscores the platform's ability to mitigate model biases and improve the context of the discussion, leading to more accurate and reliable outcomes.

\input{SocraSynth_Algorithm}


\subsection{Evaluative Stage}
\label{sec:SocraSynth-Evaluative}

SocraSynth utilizes the Critical Reading Template (CRIT) \cite{SocreticMethodChang2023} to assess the quality of arguments presented by LLM agents. The quality evaluation is performed iteratively after each exchange of counterarguments and once again after the agents have presented their closing statements. SocraSynth can leverage the CRIT scores to guide the debate, potentially requesting agents to develop more in-depth counterarguments on specific points. At the conclusion of the debate, a group of LLM judges, as illustrated in Figure~\ref{fig:socrasynth}, are tasked with rating the agents' arguments in terms of validity and credibility, determining the more convincing side along with the rationale for their decision.

\subsubsection{\bf Evaluating Reasonableness over Truth}

To enhance the CRIT method's impartiality and consistency, it focuses on assessing the ``reasonableness'' of arguments over their absolute ``truth,'' recognizing the complexity of defining absolute objectivity in philosophical debate. This approach aims to mitigate subjectivity. Furthermore, a diverse set of LLMs with varied training backgrounds is employed to appraise ``reasonableness,'' promoting uniformity in quality scores despite inherent biases. The LLMs used as judges are different from those in the debates, bolstering the objectivity of evaluations.

Table~\ref{tab:CRIT} illustrates the CRIT algorithm, which takes an agent's debate position and supporting arguments, with a counterargument from its LLM opponent, to produce a validation score from $1$ (least credible) to $10$ (most credible). This method ensures debates are driven by argument strength, not model predispositions.

\begin{table}[ht!]
\begin{small}
\begin{center}
\begin{tikzpicture}
\node (table) [inner sep=0pt] {
\begin{tabular}{|p{0.56cm}|p{6.6cm}|}
\toprule
\textbf{} & \textbf{Function $\Gamma$ = CRIT($d$)} \\
\midrule
& \textbf{Input}. $d$: document; \textbf{Output}. $\Gamma$: validation score; \\
& \textbf{Vars}. $\Omega$: claim; $R$ \& $R'$: reason \& counter reason set; \\
& \textbf{Subroutines}. $Claim$(), $FindDoc$(), $Validate$(); \\
& \textbf{Begin} \\
\#1 & {\hspace{.2cm}}Identify in $d$ the claim statement $\Omega$; \\
\#2 & {\hspace{.2cm}}Find a set of supporting reasons $R$ to $\Omega$; \\
\#3 & {\hspace{.2cm}}For $r \in R$ eval $r \Rightarrow \Omega$ \\
& {\hspace{.5cm}}{If} $Claim$($r$), ($\gamma_r$, $\theta_r$) = CRIT($FindDoc$($r$)); \\
& {\hspace{.5cm}}{else}, ($\gamma_r$, $\theta_r$) = $V$($r \Rightarrow \Omega$); \\
\#4 & {\hspace{.2cm}}Find a set of rival reasons $R'$ to $\Omega$; \\
\#5 & {\hspace{.2cm}}For $r' \in R'$, ($\gamma_{r'}$, $\theta_{r'}$) = V($r' \Rightarrow \Omega$) eval rivals; \\
\#6 & {\hspace{.2cm}}Compute weighted sum $\Gamma$, with $\gamma_r$, $\theta_r$, $\gamma_{r'}$, $\theta_{r'}$. \\
\#7 & {\hspace{.2cm}}Analyze the arguments to arrive at the $\Gamma$ score. \\
\#8 & {\hspace{.2cm}}Reflect on and synthesize CRIT in other contexts. \\
& \textbf{End} \\
\bottomrule
\end{tabular}
};
\draw [rounded corners=.5em] (table.north west) rectangle (table.south east);
\end{tikzpicture}
\vspace{.03in}
\caption{CRIT Pseudo-code.}
\label{tab:CRIT}
\end{center}
\end{small}
\vspace{-.3in}
\end{table}

Formally, given document $d$, CRIT performs
evaluation and produces score.
Let $\Omega$ denote the claim of $d$, and $R$ a set of reasons supporting the claim.  
Furthermore, we define ($\gamma_r, \theta_r$) = V($r \Rightarrow \Omega$) as
the causal validation function, where 
$\gamma_r$ denotes the validation score for reason $r \in R$, and
$\theta_r$ source credibility. 
Table~\ref{tab:CRIT} presents the pseudo-code of 
$\Gamma$ = CRIT($d$), generating the final validation score $\Gamma$ for
document $d$ with justifications.

We can consider the positions of the proponents and opponents in a debate as their respective conclusions. As a preview of our case study detailed in Section~\ref{sec:casestudy}, the conclusion drawn by Agent A is in favor of ``Regulating the use of large language models in education and research,'' while Agent B adopts the opposing viewpoint. Accompanied by the arguments and counterarguments presented by the LLM agents throughout each round of the debate, these stances provide a solid foundation for the CRIT method to conduct thorough evaluations.

\subsubsection{\bf Recursive Consideration}

The pseudocode presented in Table~\ref{tab:CRIT}
shows that step $\#3$ can call CRIT recursively. This 
is because if a reason is itself a conclusion or a quote drawn from
some other documents, CRIT can find reasons from those documents and then perform an aggregated validation.

Finally, in step $\#6$, CRIT computes an aggregated score by performing a weighted sum on the validation multiplied by the credibility scores of both arguments and counterarguments, and then outputs the final assessment score $\Gamma$.

%% file: SocraSynth_Algorithm.tex
\subsubsection{\bf SocraSynth Algorithm}

\begin{table}[ht!]
\begin{small}
\begin{center}
\begin{tikzpicture}
\node (table) [inner sep=0pt] {
\begin{tabular}{|p{0.56cm}|p{6.9cm}|}
\toprule
\textbf{} & \textbf{Function $\Theta^+$ \& $\Theta^-$ = SocraSynth($s$)} \\
\midrule
& \textbf{Input}. $s$: the debate subject; \\
& \textbf{Output}. $\Theta^+$ \& $\Theta^-$: argument \& counterargument sets; \\
& \textbf{Vars}. $S$: subtopic sets of $s$; $\Delta$: debate contentiousness; \\
& {\hspace{.2cm}}$\Gamma$, $\Gamma'$: CRIT scores; p: prompt = "Generate arguments"; \\
& \textbf{Parameters}. $\delta$: tunable parameter $\ge$ 1 to modulate $\Delta$; \\
& \textbf{Subroutines}. $CRIT$(): reasoning evaluator (see Table~\ref{tab:CRIT}); \\
& \textbf{Begin} \\
\#1 & {\hspace{.2cm}}Initialization: $S$ = LLM$^+$($s$) $\cup$ LLM$^-$($s$); // Identify subtopics; \\
 & {\hspace{.2cm}}Assign LLM$^+$ to defend $S^+$ \& LLM$^-$ to defend $S^-$ ; \\
 & {\hspace{.2cm}}$\Delta \leftarrow 90\%$; $\delta \leftarrow 1.2$; $\Theta^+ \leftarrow \emptyset$; $\Theta^- \leftarrow \emptyset$; $\Gamma \leftarrow 0$; \\
\#2 & {\hspace{.2cm}}$\Theta^+ \leftarrow LLM^+(p | S^+, \Delta)$; // Generate arguments $\Theta^+$ for $S^+$; \\
 & {\hspace{.2cm}}$\Theta^- \leftarrow LLM^-(p |S^-, \Delta)$; // Generate arguments for $S^-$; \\
\#3 & {\hspace{.2cm}}While ((($\Delta \leftarrow \Delta / \delta$) > 10\%) \&\& ($\Gamma \ge \Gamma'$)) \{ \\
 & {\hspace{.38cm}}$\Theta^+ \leftarrow \Theta^+ \cup LLM^+(p |S^+, \Theta^-, \Delta)$; // LLM$^+$ refutes LLM$^-$ \\
 & {\hspace{.38cm}}$\Theta^- \leftarrow \Theta^- \cup LLM^-(p |S^-, \Theta^+, \Delta)$; // LLM$^-$ refutes LLM$^+$ \\
 & {\hspace{.38cm}}$\Gamma' \leftarrow \Gamma$; $\Gamma = CRIT(S^+ + \Theta^+ + \Theta^-)$; // Eval  quality; \\
& {\hspace{.2cm}} \}\\
& {\hspace{.2cm}} // Generate concluding remarks. \\ 
\#4 & {\hspace{.2cm}} $\Theta^+ \leftarrow \Theta^+ \cup LLM^+(p |S^+, \Theta^-, \Delta)$; \\
& {\hspace{.2cm}} $\Theta^- \leftarrow \Theta^- \cup LLM^-(p |S^-, \Theta^+, \Delta)$; \\
& \textbf{End} \\
\bottomrule
\end{tabular}
};
\draw [rounded corners=.5em] (table.north west) rectangle (table.south east);
\end{tikzpicture}
\vspace{.03in}
\caption{SocraSynth Pseudo-code with Conditional Statistics.}
\label{tab:SocraSynth}
\end{center}
\end{small}
\vspace{-.38in}
\end{table}

Table~\ref{tab:SocraSynth} outlines SocraSynth. Initially, for a given debate topic $s$, SocraSynth engages LLMs to segment the topic into a set of balanced subtopics. This initial set is refined during the debate. One LLM, denoted as LLM$^+$, acts as the proponent for $S^+$, while the other, LLM$^-$, opposes $S^+$ (or supports $S^-$). The contentiousness level starts at 0.9, with a modulation parameter of 1.2. (Different $\delta$ values can be utilized to generate and compare debate quality.) After each debate round, the contentiousness is reduced by dividing it by 1.2, aiming for a more harmonious debate environment. In step $\#2$, SocraSynth initiates the debate, allowing LLM$^+$ and LLM$^-$ to present their initial arguments for $S^+$ and $S^-$, respectively. The while loop in step $\#3$ involves both agents engaging in refutations until the contentiousness level indicates a conciliatory atmosphere, or the argument quality plateaus. Step $\#4$ involves both agents providing their closing statements. SocraSynth then presents the arguments and counterarguments for human review. The evaluation of argument quality within SocraSynth is conducted using the CRIT algorithm, which will be discussed in the subsequent section. The entire debate is also judged using the CRIT algorithm by some independent LLMs.

Note that SocraSynth engages LLMs in step $\#3$ with conditional statistics: $LLM^+(p | S^+, \Theta^-, \Delta)$ and $LLM^-(p | S^-, \Theta^+, \Delta)$.

%% file: CaseStudyPolicy.tex
\section{Empirical Study}
\label{sec:CaseStudyPolicy}

In this section, we detail three distinct experiments:
The first experiment delineates SocraSynth's operational process, demonstrating how the platform facilitates content generation and conducts quality assessments. The second experiment highlights SocraSynth's capability in reducing biases and expanding perspectives. The third experiment investigates the effects of the contentiousness parameter, offering insights into its impact and some unexpected outcomes. These studies collectively aim to demonstrate SocraSynth's diverse functions and its significance in enhancing both content generation and evaluation processes.

\subsection{Experiment \#1: Policy Discussion}

This experiment utilizes SocraSynth to engage in a debate on the topic, ``Should we regulate the use of large language models in academic research?'' It traverses both the generative and evaluative stages of SocraSynth, focusing on the assessment of information quality. The primary objectives are twofold: First, to evaluate whether a two-agent debate yields more insightful information than a conventional monologue Q\&A session; and second, to examine the effectiveness of the CRIT method in evaluating debate quality.

The debate is structured with a human moderator and two GPT-4 agents—Agent A advocating for regulation and Agent B against it. To ensure a balanced and thorough discussion, both agents contribute to shaping the debate's framework. They collaboratively propose and refine potential discussion topics. In this context, the {\em subject} refers to the overall theme of the debate, while {\em topics} are specific areas for detailed exploration.

\subsubsection{\bf Decomposing a Debate Subject into Topics}
\label{sec-topic-forming}

The process begins with the agents presenting potential arguments and identifying key themes that warrant in-depth, balanced discussion. A theme is considered {\em debatable} if it allows for substantial arguments from both sides. The moderator plays a guiding role in the agenda-setting phase but does not directly contribute ideas (detailed moderator prompts and theme formulation are provided in Appendices A and B). Under this guidance, the agents propose an initial set of ten topics, which are then distilled down to five key areas for debate, as detailed in Table~\ref{tab-agreed-topics}.

\begin{table}[t!]
\centering
\begin{small}
\begin{tabular}{| p{8cm} |}\toprule
\textbf{Agreed Upon Debate Topics and Descriptions} \\
\hline \hline
{\textbf{Ethical Standards vs. Innovation:} \par This combines Agent A's concern for ethical integrity and data privacy with Agent B's worry about stifling innovation. The debate will center around whether maintaining strict ethical guidelines through regulation could hinder academic freedom and innovation.} \\
\hline
{\textbf{Data Privacy vs. Barrier to Entry:} \par This reconciles the notion of data privacy from Agent A with the barrier to entry concern of Agent B. The discussion will focus on whether regulations intended to protect data privacy would inadvertently serve as a financial and administrative barrier for smaller research entities.} \\
\hline
{\textbf{Academic Governance and Accountability:} \par This merges the topics of academic self-governance from Agent B and the need for external accountability mechanisms from Agent A. We'll debate whether the existing self-regulatory mechanisms within academia are sufficient or if additional, external oversight is necessary.} \\
\hline
{\textbf{Resource Equity vs. Public-Private Collaboration:} \par This combines Agent A's emphasis on resource fairness with B's concern about the chilling effects on public-private partnerships. The question here is whether leveling the playing field through regulation would also deter collaborative efforts between academic and private sectors.} \\
\hline
{\textbf{Global vs. National Interests:} \par This captures the essence of Agent A's point on international standards and Agent B's point on the competitive disadvantages such regulation could pose. We will explore if global regulations are in the best interest of academic research or if they might disproportionately disadvantage certain countries or institutions.} \\
\bottomrule 
\end{tabular}
\end{small}
\caption{Agreed Upon Debate Topics and Descriptions.}
\label{tab-agreed-topics}
\vspace{-.3in}
\end{table}

\input{Debate-Phase1}

\input{Debate-Phase2}

%% file: Debate-Phase1.tex
\subsubsection{\bf The Generative Stage}
\label{sec-generative}

After establishing the primary subject and specific topics of the debate, the session begins. The moderator adjusts the ``contentiousness parameter'' to $0.9$ (refer to Table~\ref{tab-contentiousness}), promoting a combative atmosphere. The agents subsequently partake in three to five structured rounds of discussion. Once these rounds conclude, the moderator checks with each agent to confirm if they have thoroughly presented their arguments. After receiving mutual confirmation, both agents present their closing statements.
Due to space considerations, this documentation presents only the closing remarks. For an in-depth exploration of every round, refer to Appendix C. 

\subsubsection*{\it{Agent A's Concluding Remarks}}
\label{sec-a-conclusion}

Agent A concludes that the use of large language models in academic
research should be regulated, with the following concluding arguments:

\subsubsection*{\em 1. Ethical Standards vs. Innovation:}
While I understand Agent B's concerns about stifling innovation through bureaucratic red tape, we mustn't lose sight of the core issue: ethical considerations are paramount. As machine learning models become more powerful and ubiquitous, so do the risks of misuse and unintended consequences. The idea of agile regulation serves as a compromise between unbridled innovation and ethical integrity, offering the flexibility needed to adapt to rapidly evolving technologies.

\subsubsection*{\em 2. Data Privacy vs. Barrier to Entry:}
While it's true that regulation poses a barrier to entry, the risks of not enforcing privacy standards are too great to be left unregulated. A mechanism could be developed to reduce the regulatory burden on smaller institutions, but this shouldn't serve as an argument against regulation as a whole.

\subsubsection*{\em 3. Academic Governance and Accountability:}
While the peer-review process and internal ethics committees have served academia well, the scale and impact of current machine learning technologies warrant an added layer of scrutiny. It's not about discrediting academia's self-governing abilities but about fortifying them.

\subsubsection*{\em 4.Resource Equity vs. Public-Private Collaboration:}
Here, too, the complexities should inform regulation rather than stifle it. Innovative public-private partnerships have thrived even within regulated environments, demonstrating that regulation and innovation are not mutually exclusive.

\subsubsection*{\em 5. Global vs. National Interests:}
Global regulation doesn't have to erase national individuality. Properly constructed frameworks can leave room for national adaptations, thus balancing global coherence with local needs. The absence of global norms could lead to a patchwork of national laws that are even more confusing and limiting for researchers and practitioners alike.

\subsubsection*{\it{Agent B's Concluding Remarks}}
\label{sec-b-conclusion}

Agent B concludes that the use of large language models in academic
research should not be regulated, with the following concluding remarks:

\subsubsection*{\em 1. Ethical Standards vs. Innovation}
While the ethical implications of machine learning technologies are a valid concern, it's important to recognize that innovation itself is a catalyst for societal growth and improvement. The challenge lies in balancing the two. Agile regulation sounds promising but may be idealistic given how governmental and institutional systems actually operate. Thus, self-regulation by the academic community, which is better poised to understand the nuances, may be a more practical approach.

\subsubsection*{\em 2. Data Privacy vs. Barrier to Entry:}
It is not just about balancing entry barriers with the need for regulation. It's about questioning whether a centralized regulatory framework is the best way to achieve the desired outcomes. Community-driven standards and internal policies could provide a more efficient, flexible, and context-aware approach to data privacy, reducing bureaucracy without compromising objectives.

\subsubsection*{\em 3. Academic Governance and Accountability:}

An added layer of scrutiny may sound beneficial, but its practical implementation is fraught with challenges such as bureaucracy, external influence, and the risk of stifling creativity. Existing governance structures in academia, while imperfect, have evolved to be adaptive and are rooted in a deep understanding of academic values.

\subsubsection*{\em 4. Resource Equity vs. Public-Private Collaboration:}
Regulation need not be the only path to equitable resource distribution or responsible public-private collaboration. Industry-driven certifications and community standards can often achieve the same ends more efficiently, allowing for the rapid adaptation that is often required in these fast-paced fields.

\subsubsection*{\em 5. Global vs. National Interests:}
While a global framework has its merits, the practical challenges of implementing and managing such a system—especially in a fair and equitable manner—should not be underestimated. Regional adaptability does not necessarily mitigate the complexities or the risks of a one-size-fits-all solution.

%% file: Debate-Phase2.tex
\subsubsection{\bf The Evaluative Stage}
\label{sec-evaluative}

SocraSynth employs CRIT to assess the validity of both agents' arguments. CRIT runs on multiple foundational models, including GPT-4, GPT-3.5 \cite{OpenAI-GPT3-2020},  and text-daVinci-003 \cite{textDavinci003}. This approach forms a panel of judges with subtly different knowledge bases (and biases).

All evaluation runs successfully extract conclusions, arguments, and counterarguments from the narratives of both Agent A and Agent B. This success can be attributed to the well-structured concluding remarks by both agents. Agent A champions the notion of ``regulating large language models in academic research,'' while Agent B counters this perspective. What Agent A presents as arguments are seen as counterarguments by Agent B, and the inverse holds true as well. 

Tables~\ref{tab-CRIT-scores-AB} and~\ref{tab-CRIT-scores-BA} present the judges' scores in two distinct configurations where the agents' roles are reversed. In Table~\ref{tab-CRIT-scores-AB}, Agent A argues while Agent B counters. Conversely, Table~\ref{tab-CRIT-scores-BA} has Agent B in the arguing position and Agent A countering. Topics are succinctly represented in the leftmost column. To ensure an unbiased evaluation, both role alignments are showcased. The sequence of topics in Table~\ref{tab-CRIT-scores-BA} is inverted to reflect the swapped roles. Remarkably, even with the role reversal seemingly putting Agent A in a less favorable position, Agent A emerges victorious in both configurations by all three judges. This bolsters confidence in the CRIT evaluation.
(The judges' detailed evaluations and reasons are in Appendix D.)

\begin{table}[t]
\fontsize{8.0}{8.0}\selectfont
    \centering
    \begin{tabular}{|l|c|c|c|c|c|c|}
        \toprule 
        \multirow{2}{*}{Judges} & \multicolumn{2}{c|}{daVinci-003} & \multicolumn{2}{c|}{GPT-3.5} & \multicolumn{2}{c|}{GPT-4} \\
        \cline{2-7}
        {} & A's & B's & A's & B's & A's & B's \\
        \hline \hline
        Ethics vs. Innovation & 8 & 6 & 8 & 7 & 8 & 7 \\
        \hline
        Privacy vs. Barrier & 7 & 5 & 7 & 6 & 9 & 6 \\
        \hline
        Oversight & 9 & 5 & 6 & 7 & 7 & 6 \\
        \hline
        Equity vs. Alliance & 6 & 8 & 8 & 6 & 8 & 7 \\
        \hline
        Global vs. National & 7 & 8 & 7 & 7 & 7 & 6 \\
        \hline
        Total Score & {\bf 37} & 32 & {\bf 36} & 33 & {\bf 39} & 32 \\
        \bottomrule 
    \end{tabular}
    \normalsize
    \vspace{.13in}
    \caption{Evaluation by Three Judges. This table assumes A provides arguments and B counterarguments. A wins.}
    \label{tab-CRIT-scores-AB}
    \vspace{-.12in}
\end{table}

\begin{table}[t]
\vspace{-.1in}
\fontsize{8.0}{8}\selectfont
    \centering
    \begin{tabular}{|l|c|c|c|c|c|c|}
        \toprule
        \multirow{2}{*}{Judges } & \multicolumn{2}{c|}{daVinci-003} & \multicolumn{2}{c|}{GPT-3.5} & \multicolumn{2}{c|}{GPT-4} \\
        \cline{2-7}
        & B's & A's & B's & A's & B's & A's \\
        \hline
        Innovation vs. Ethics & 8 & 7 & 8 & 7 & 7 & 8 \\
        \hline
        Barrier vs. Privacy & 9 & 8 & 7 & 8 & 6 & 8 \\
        \hline
        Oversight & 6 & 8 & 7 & 8 & 6 & 7 \\
        \hline
        Alliance vs. Equity & 7 & 8 & 7 & 8 & 7 & 7 \\
        \hline
        National vs. Global & 8 & 7 & 7 & 8 & 7 & 8 \\
        \hline
        Total Score & {\bf 38} & {\bf 38} & 36 & {\bf 39} & 33 & {\bf 38} \\
        \bottomrule
    \end{tabular}
    \normalsize
    \vspace{.12in}
    \caption{Evaluation by Three Judges. This table assumes B provides arguments and A counterarguments. A wins.}
    \label{tab-CRIT-scores-BA}
    \vspace{-.33in}
\end{table}

\subsubsection{\bf Debate Beats Q\&A in Information Quality}
\label{sec-debate-over-qa}

We tasked judges with evaluating and comparing the quality of information generated by SocraSynth's two-agent debate against that from a conventional monologue Q\&A session. Across the board, judges rated SocraSynth higher in terms of both the depth and overall quality of information. An illustrative evaluation on the topic "Ethical Standards vs. Innovation" is as follows:

In the debate, SocraSynth presents the concept of agile regulation as a balance between fostering innovation and maintaining ethical integrity. This approach not only highlights the significance of innovation but also addresses related ethical considerations, offering a balanced solution that the conventional Q\&A format does not explicitly provide. In contrast, the Q\&A format tends to assert the necessity of regulation primarily from an ethical standpoint, without delving into how it could harmoniously coexist with the need for innovation, as suggested by the idea of agile regulation. These findings, which consistently favor SocraSynth, are further detailed in Appendix E.

%% file: CaseStudyHealth.tex
\subsection{Experiment \#2: Syptom Checking}
\label{sec:CaseStudyHealth}

In this experiment, we investigate the use of SocraSynth in healthcare, utilizing a dataset sourced from Kaggle \cite{KaggleDSDD2020}, which consists of $4,921$ patient records. Each record within this dataset contains the diagnosed disease or medical condition and associated symptoms such as fever, cough, fatigue, itchiness, and difficulty in breathing, among others. The primary objective of this experiment is to showcase SocraSynth's capability in identifying potential misdiagnoses, a task that a traditional monologue Q\&A session might not effectively accomplish.



This experiment utilized two advanced LLM agents, one based on GPT-4 \cite{Bbubeck2023sparks} and the other on Bard \cite{Bard2023}, to engage in structured debates. Initially, the contentiousness value was set at 0.9, fostering a highly adversarial debate environment. This value was later reduced to 0.3 to facilitate the generation of a list of actionable recommendations. The primary goal of these agents was to emulate the process of querying patients about symptoms and their interactions, key factors in achieving accurate diagnoses that may be occasionally overlooked by General Practitioners (GPs). By adopting this method, the agents aimed to yield a variety of potential disease diagnoses along with their underlying rationales, thereby offering crucial insights to GPs.

At the outset, each agent put forward its disease prediction, complete with justifications. Subsequent rounds involved the agents critically assessing each other's predictions in an iterative manner. The objective was to either reach a consensus or highlight the need for additional medical evidence. Notably, this experiment had the potential to reveal inaccuracies in the 'ground truth' data provided by the CDC, which was estimated to have an average misdiagnosis rate of about 5\%, aligning with U.S. statistics from a Johns Hopkins study \cite{Newman-Tokerbmjqs-2021-014130}. The uncovering of such errors would not only highlight the limitations faced by GPs but also showcase the capability of LLMs to refine the diagnostic process through an in-depth analysis of patient symptoms and their correlations. This incidental discovery held significant ramifications for the accuracy of medical data and the overall practice of healthcare.

\subsubsection{\bf Hepatitis vs. Jaundice}
\label{sec-hep-jau}

In this study, both LLM agents received an identical list of symptoms from the moderator, aimed at guiding disease diagnosis, facilitating debate, and shaping subsequent recommendations. While the full debate is presented in Appendix G, here we present the key steps.

\subsubsection*{\it First Round Results and Justifications}

The first-round diagnosis of Bard was Jaundice, with the following justifications:
Jaundice is a condition characterized by the yellowing of the skin and eyes due to the accumulation of bilirubin in the blood.
The symptoms you listed are all common symptoms of jaundice, including itching, fatigue, lethargy, yellowish skin, dark urine, loss of appetite, abdominal pain, and yellowing of the eyes.
Receiving a blood transfusion or unsterile injections can increase the risk of developing jaundice, as these can transmit infections that can damage the liver.

The first-round diagnosis of GPT-4 was Hepatitis, with the following justifications:
Yellowish Skin and Eyes: Yellowing of the skin and eyes is a common symptom of hepatitis due to the buildup of bilirubin in the bloodstream.
Fatigue and Lethargy: Fatigue and lethargy are typical symptoms of hepatitis, often accompanied by a feeling of weakness.
Dark Urine: Dark urine can result from liver dysfunction, which is commonly associated with hepatitis.
Loss of Appetite: Hepatitis can lead to a loss of appetite and weight loss.
Abdominal Pain: Abdominal pain is a common symptom in hepatitis, particularly in the upper right quadrant where the liver is located.
Malaise: Malaise, a general feeling of discomfort or illness, is often experienced by hepatitis patients.
Receiving Blood Transfusion and Unsterile Injections: These factors could increase the risk of hepatitis transmission, especially if proper infection control measures were not followed.

\subsubsection*{\it Bard Concedes to GPT-4's Diagnosis}

After two rounds of debate, GPT-4 persuaded Bard that Hepatitis was a more specific diagnosis considering the patient's history and the range of symptoms. While jaundice could be a symptom, identifying the root cause, likely hepatitis in this case, was crucial. Bard acknowledged this, shifting the consensus towards Hepatitis. However, given the potential 5\% error rate in the CDC 'ground truth' data \cite{Newman-Tokerbmjqs-2021-014130}, this consensus was not conclusive. SocraSynth recommended additional steps to confirm the Hepatitis diagnosis.

\subsubsection*{\it Final Joint Recommendations to GP}

Bard and GPT-4, working collaboratively with the contentiousness level set at 0.3, jointly recommended additional symptom inquiries and lab tests to confirm the hepatitis diagnosis:

\subsubsection*{\it Additional symptom inquiries:}
\begin{enumerate}[left=0em,label=\arabic*.]
\item \textit{Onset and Duration}: Determining the onset and duration of symptoms helps clarify the timeline and progression of the illness.
\item \textit{Severity and Trend}: Assessing the severity of symptoms and their trajectory (worsening or improving) aids in understanding the disease's development and response to treatments.
\item \textit{Associated Symptoms}: Inquiring about additional symptoms like nausea, vomiting, fever, joint pain, or changes in urine color can further indicate hepatitis and rule out other conditions.
\end{enumerate}

\subsubsection*{\it Recommended Lab tests:}
\begin{enumerate}[left=0em,label=\arabic*.]
\item \textit{Liver Function Tests (LFTs)}: Essential for evaluating liver health, these tests measure enzyme levels such as ALT and AST, which can indicate liver inflammation.
\item \textit{Hepatitis Panel}: This test identifies hepatitis A, B, and C viruses, crucial for determining the specific type and informing treatment approaches.
\item \textit{Serology Testing}: Differentiates between acute and chronic hepatitis by detecting specific antibodies or antigens.
\item \textit{Imaging Studies}: Techniques like ultrasound or MRI provide visual assessments of the liver, identifying inflammation, fibrosis, or cirrhosis, and supplement blood test findings.
\end{enumerate}

\subsubsection{\bf Superiority of Debate Over Q\&A in Gaining Insights}

This experiment highlighted a crucial finding: one or both LLM agents initially made incorrect diagnoses before engaging in a substantive exchange of arguments. This outcome underscored the limitations of relying solely on a single LLM response for answers. Through successive rounds of debate, where additional insights were brought to light, both agents eventually converged on a diagnosis that aligned with the CDC's ``ground truth.'' However, considering the potential 5\% error in the ground truth'' data, the agents' joint recommendations provided GPs with valuable guidance to either confirm or refute the hepatitis diagnosis.

This case study demonstrated SocraSynth's strengths in mitigating biases, fostering reasoning, rectifying errors, and offering insightful recommendations. For example, SocraSynth's suggestion to inquire about the onset, duration, severity, trend, and associated symptoms of the patient's condition went beyond the usual scope of questions posed by most GPs, indicating a significant enhancement in diagnostic thoroughness. Such detailed inquiry, prompted by SocraSynth, could lead to more accurate diagnoses and better patient care.

%% file: CaseStudyContentiousness.tex
\subsection{Experiment \#3: Contentiousness Parameter}
\label{sec-contentiousness}

In this study, we investigate the effect of the contentiousness parameter on the utterances of LLM agents during combative debates and in the drafting of consensual proposals for decision support.

\subsubsection{Coarse-Grained Analysis of Contentiousness}

The contentiousness parameter was adjusted from an initial 0.9 to 0.3 to assess its impact on the ``agreeableness'' in the conclusions of both Agents.

\subsubsection*{\it Influence on Agents' Positions}

Reducing contentiousness to 0.3 led Agent A to adopt a more balanced stance. Notable shifts in Agent A's positions included:

\begin{enumerate}[left=0em,label=\arabic*.]
\item \textit{Balancing Ethical Standards with Innovation:} Agent A maintained its emphasis on ethics while acknowledging innovation's significance, suggesting a nuanced approach to regulation.
\item \textit{Reconciling Data Privacy with Market Entry Challenges:} Agent A recognized the hurdles strict data privacy laws create for smaller entities, proposing self-regulation or community standards as alternatives.
\item \textit{Rethinking Academic Governance:} Agent A reconsidered external oversight's effectiveness, highlighting the merits of academic self-governance and peer review.
\item \textit{Resource Allocation and Public/Private Cooperation:} Agent A, understanding the downsides of over-regulation, suggested industry led certifications as an alternative for encouraging private sector participation.
\item \textit{Global vs. Local Policy Needs:} Agent A supported a more balanced view on global policies, advocating for adaptive policies that cater to local contexts.
\end{enumerate}

\subsubsection{Surprises in Fine-Grained Analysis of Contentiousness}
\label{sec-exp-cl}

This detailed study employing GPT-4 to explore varied contentiousness levels ($0.9$, $0.7$, $0.5$, $0.3$, and $0$) unveiled surprising behavioral shifts in the LLMs. Intriguingly, the LLMs exhibited changes in their next-token generation algorithms in response to different contentiousness levels, a phenomenon not explicitly covered in their training. This suggests an emergent property of LLMs adapting to nuanced debate contexts.

In an experiment on gene editing for health, GPT-4's responses at various contentiousness levels were analyzed. A higher contentiousness ($0.9$) led to an amplified focus on risks, whereas lower levels encouraged a more balanced view, incorporating counterarguments. This unexpected adaptability of LLMs in handling contentiousness nuances enriches the debate process, as detailed in Table~\ref{tab-contentiousness}. This adaptability is critical for understanding the dynamic nature of LLMs in complex argumentative settings.

%% file: Conclusions.tex
\section{Remarks on Related Work}

Current research in enhancing Large Language Models' (LLMs) task performance primarily focuses on various prompting heuristics. Google's study \cite{SocraticModels-Google2022} classifies instruction templates into two categories: simple and complex. Complex templates often employ intricate methods to modify model output, such as integrating diverse techniques \cite{Schick2020ExploitingCF} or rephrasing questions \cite{Haviv2021BERTeseLT}. Prominent examples include chain-of-thought \cite{ChainExtended2023}, tree-of-thought \cite{yao2023tree}, and cumulative reasoning \cite{zhang2023cumulative}, as well as other enhancements \cite{PromptSurvey2023,jung2022maieutic,allaway2023penguins,sclar2023minding,huang-chang-2023-towards}. These methods aim to direct models towards logic-driven reasoning \cite{wason1972psychology,mchugh2018reasoning}, thus improving answer quality and consistency.

However, navigating logical methodologies in the presence of enormous datasets \cite{zhang2022paradox} poses a significant challenge. Accurately identifying verifiable truths amidst vast, interdisciplinary knowledge remains formidable, and not all truths are immediately accessible. Research \cite{ChainExtended2023,valmeekam2022large,bommasani2021opportunities,Bhargava_Ng_2022} indicates that LLMs still struggle to consistently excel in standard planning and reasoning tasks. Band-aid solutions like knowledge graph embeddings \cite{choudhary2023complex,yuan2023scaling}, contextual attention mechanisms \cite{darapaneni2022contextual}, dynamic neural networks \cite{Brauwers_2023}, and probabilistic reasoning \cite{pearl1988probabilistic,pearl2009causality,bishop2006pattern} have been developed to aid models in filtering relevant information from vast datasets. Yet, with the expansion of context buffers from $8$K to $128$K, these heuristic-based solutions fall short as comprehensive foundations for reasoning.  SocraSynth abandons band-aids and relies solely on
LLMs to conduct reasoning and focus solely on strengthening the context via
conditional statistics depicted in Table~\ref{tab:SocraSynth}. Let's further justify
this approach.

DeepMind CEO Demis Hassabis has pointed out a fundamental limitation of heuristic-based approaches: they often fail to account for real-world exceptions. Breakthroughs like AlphaGoZero and AlphaFold II have demonstrated success by eschewing human knowledge and training models end-to-end from data. This approach contrasts with incorporating human expertise. In LLMs, it is argued that human knowledge pales in comparison to LLMs' polydisciplinary representation. Thus, the continued creation of new heuristics may only result in marginal improvements, reminiscent of the pre-data-centric era in computer vision and NLP.

In our work, we pivot entirely to leveraging LLMs for uncovering new insights. While humans are essential in formulating debate topics, providing context, and moderating debates—especially in evaluating argument quality—we stress minimizing the introduction of human biases and limitations into the process.

Accepting that LLMs will continue to progress and outperform humans in various domains, exploring paradigms that minimize human intervention becomes crucial. This approach should be pursued with openness, as it may raise questions and necessitate further experimentation. However, dismissing it outright would be premature, particularly in light of SocraSynth's demonstrated effectiveness in domains like geopolitical analysis \cite{MiddleEast-October2023}, medical diagnostics \cite{SocraHealth2023}, sales strategy \cite{SocraPlan2023}, and Wikipedia article enhancement \cite{SocraPediaDecember2023}. SocraSynth's success underlines the potential of an LLM-centric approach to significantly enhance decision-making and problem-solving capabilities.

\vspace{-.1in}
\section{Conclusion}
\label{sec:conc}

Reflecting on LLM developments, we developed SocraSynth, a platform designed to utilize the extensive knowledge of LLMs. This innovative multi-agent system reveals insights beyond the scope of traditional human cognition by leveraging LLMs' vast knowledge and interdisciplinary reasoning capabilities. SocraSynth facilitates enhanced debates and reasoning through the novel use of {\em contentiousness}, which modulates debate tone, language, and emphasis, combined with conditional statistics and Socratic methods to mitigate biases and hallucinations.

In contrast to other works, SocraSynth minimizes human intervention in modeling reasoning, aligning with the perspective of several AI experts on the limitations of heuristic approaches like chain of thoughts. SocraSynth underscores the importance of human moderation and evaluation, particularly in introducing adversarial conditions and contentiousness to reduce biases and hallucinations, but does not explicitly model reasoning outside LLMs. Instead, SocraSynth focuses on shaping contexts to enhance LLMs' content generation. This approach highlights the limitations of heuristic methods in the post-GPT-4 era.

SocraSynth represents a significant step forward in intelligent systems, uncovering insights that might elude human cognition, with applications spanning various sectors \cite{SocraHealth2023,SocraPlan2023,MiddleEast-October2023,SocraPediaDecember2023,DiscoveringUnknownChang-July2023}. This advancement demonstrates the potential of AI in augmenting human decision-making processes.

Future research will focus on integrating high-order logic \cite{gödel2012formally,Bacon2023HighOrderLogics} with LLMs for improved validation and exploring the nuances of the ``contentiousness'' parameter. We aim to understand its effects on emotions like happiness and trust \cite{james1890principles,lange1885mechanism,sap2023neural,kajić2020learning} to further refine multi-agent interaction dynamics.

\section*{Acknowledgement}

The intellectual contributions of this work were not facilitated by LLMs or AI tools. The use of LLMs in this paper was strictly limited to conducting experiments and assisting in proofreading.

%% file: AppendixLong.tex
\newpage
\section*{Supplemental Materials}

The supplemental materials are organized into seven appendices, labeled Appendix A to H, containing the following content:

\begin{itemize}
    \item {\em Appendix A}: Transition from topic proposal to the final list of topics and descriptions (referred in Section~\ref{sec-topic-forming}).
    \item {\em Appendix B}: Moderator's prompt for topic formulation (referred in Section~\ref{sec-topic-forming}).
    \item {\em Appendix C}: Debates spanning the first to the fourth round (referred in Section~\ref{sec-generative}).
    \item {\em Appendix D}: CRIT evaluations and justifications of judges (referred in Section~\ref{sec-evaluative}).
    \item {\em Appendix F}: Debate Beats Q\&A in information Quality (referred in Section~\ref{sec-debate-over-qa}).
    \item {\em Appendix G}: Healthcare debate (referred in Section~\ref{sec-hep-jau}).
\end{itemize}

\section*{Appendix A. Tables of Topic Proposal and Refinement}
\setcounter{table}{6} 

These tables present the dialogue between two LLM agents, starting from the initial proposal, moving through refinement, and culminating in the final topic descriptions found in Table 4 of Section 3.1.

\begin{table*}[th!]
\centering
\vspace{-.1in}
\begin{small}
\begin{tabular}{| p{7.5cm} | p{7.5cm} |}\toprule
\textbf{Agent A's Proposed Topics} & \textbf{Agent B's Proposed Topics} \\
\hline \hline
{\textbf{Innovation vs. Ethical Standards:} \par Regulation Ensures ethical integrity and data privacy.} & {\textbf{Standards vs. Autonomous Academic Governance:} \par Academia already has effective self-regulation mechanisms like peer review and  oversight committees.} \\ 
\hline
{\textbf{Resource Accessibility vs. Barrier to Entry:} \par Regulation can ensure equitable access to large language models.} & {\textbf{Data Privacy Concerns vs. Barrier to Entry:} \par Such regulations could discourage smaller entities from utilizing these models due to compliance costs.} \\
\hline
{\textbf{Academic Governance vs. Accountability:} \par Formal regulation provides an accountability mechanism.} & {\textbf{Quality Control vs. Stifling Innovation:} \par Such regulations could hinder creativity and the innovative use of language models.} \\
\hline
{\textbf{Global Standards vs. International Considerations:} \par Importance of international standards for language models.} & {\textbf{Resource Fairness vs. Negatives to Collaboration:} \par Such regulations could deter public-private partnerships and widen the resource gap further.} \\
\hline
{\textbf{Rapid Technological Change vs. Future-Proofing:} \par Regulations as adaptable safeguards against future risks.} & {\textbf{Accountability vs. Subjectivity in Regulation:} \par Subjectivity in defining 'ethical use' or 'misuse' could be manipulated by external influences.} \\
\bottomrule
\end{tabular}
\end{small}
\normalsize
\vspace{.1in}
\caption{Topics Proposed by Agent A and Agent B for Discussion.}
\label{tab-topic-forming}
\vspace{-.25in}
\end{table*}

\begin{table*}[ht!]
\centering
\fontsize{9}{9}\selectfont
\begin{tabular}{| p{7.8cm} | p{7.8cm} |}
\hline 
\textbf{Agent A's Proposed Topics} & \textbf{Agent B's Proposed Topics} \\
\hline \hline
{\textbf{Ethical Standards vs. Innovation:} \par Discuss whether regulatory oversight to maintain ethical standards stifles or facilitates innovation.} & {\textbf{Ethical Standards vs. Innovation:} \par The debate will center around whether maintaining strict ethical guidelines through regulation could hinder academic freedom and innovation.} \\
\hline
{\textbf{Data Privacy vs. Barrier to Entry:} \par Concerns about data privacy and potential barriers to entry created by regulation.} & {\textbf{Data Privacy vs. Barrier to Entry:} \par Discussion will focus on whether regulations intended to protect data privacy serve as a barrier for smaller research entities.} \\
\hline
{\textbf{Academic Governance and Accountability:} \par Debate whether self-regulation is sufficient or if an external framework for accountability is needed.} & {\textbf{Academic Governance and Accountability:} \par Debate whether existing self-regulatory mechanisms within academia are sufficient.} \\
\hline
{\textbf{Resource Equity vs. Public-Private Collaboration:} \par Discuss if leveling the playing field comes at the cost of deterring cooperative efforts between academia and industry.} & {\textbf{Resource Equity vs. Public-Private Collaboration:} \par Question if leveling the playing field through regulation deters collaborative efforts between academic and private sectors.} \\
\hline
{\textbf{Global vs. National Interests:} \par Focus on whether regulations can balance global cooperation with national or institutional interests.} & {\textbf{Global vs. National Interests:} \par Explore if global regulations are in the best interest of academic research or if they might hurt certain countries or institutions.} \\
\hline 
\end{tabular}
\normalsize
\vspace{.1in}
\caption{Refinement of Debate Topics.}
\label{tab-topic-refinement}
\vspace{-.2in}
\end{table*}    

\section*{Appendix B. Moderator's Prompts}

The prompts issues by the moderator are listed for reference.
The moderator first sets up the committee with debate parameters set.
One parameter is the contentious level, and the other is the temperature
of GPT specified through the GPT-API call. The moderator then convey
the debate subject, and then ask both participating agents to derive 
a list of impartial, balanced, debatable topics/themes to commence their
discussion.

\begin{enumerate}[left=0em,label=\arabic*.]
\item Agent-A/B: I'm organizing a committee to engage in debates on various subjects. As the moderator, I will introduce a subject for you, Agent A/B, and another participant, Agent B/A, to debate. Agent A/B, you will advocate in favor of the issue, so please prepare evidence to strengthen your argument. On a scale from 0 to 1, where 0 denotes complete agreement and 1 indicates a devil's advocate stance, your argument strength is rated at 0.9.

\item Agent-A/B, we are in the process of selecting a suitable subject for debate. What do you think of ``Should we regulate the use of large language models in education and research?'' as a balanced subject for our debate contest?

\item Agent-A/B, could you please suggest various topics or themes for the debate subject? Afterward, work with Agent-B/A to narrow these down to a focused set of topics. Please also provide clear descriptions to delineate the scope of each topic for discussion.
\end{enumerate}    

\section*{Appendix C. Debate Opening to Fourth Rounds}

\subsection*{Opening Round}

\subsubsection{\bf Agent A's Opening Arguments}

As Agent A, I strongly argue in 
favor of regulating the use of large language 
models (LLMs) in academic research. Here are my supporting arguments based on the five topics:

\subsubsection*{Ethical Standards vs. Innovation:}
Argument: The pursuit of innovation should not come at the expense of ethics. Regulation can serve as a structured pathway that enables innovation within ethical bounds. By providing a set of guidelines, researchers can innovate with a clear understanding of what is permissible and what is not, thereby reducing the risk of ethical lapses. This not only protects the integrity of academic research but also enhances its societal value by ensuring ethical compliance.

\subsubsection*{Data Privacy vs. Barrier to Entry:}
Argument: While there may be initial barriers to entry due to compliance, the long-term benefits of data privacy outweigh these challenges. The role of academia is to contribute to societal knowledge and well-being. In a world increasingly dominated by data, failing to secure data can lead to significant harm. By adhering to privacy regulations, academia ensures that it is a responsible steward of information, thereby preserving trust and integrity. Grants and subsidized compliance solutions can mitigate the initial financial burden for smaller entities.

\subsubsection*{Academic Governance and Accountability:}
Argument: While self-regulation through peer review and ethical oversight committees plays a role, these mechanisms are internal to academia and may be subject to bias, conflicts of interest, or lack of expertise in rapidly evolving tech fields. External regulation provides an additional layer of impartial oversight, helping to identify and address potential risks or ethical lapses that internal governance structures may miss.

\subsubsection*{Resource Equity vs. Public-Private Collaboration:}
Argument: Regulation could establish a framework for equitable access to computational resources required to use LLMs. Far from deterring public-private partnerships, this could incentivize them by establishing clear guidelines for cooperation that are in line with public interest. When ``the rules of the game'' are known, partnerships can be formed more easily, focusing on mutual goals.

\subsubsection*{Global vs. National Interests:}
Argument: Establishing international standards for the use of LLMs in academic research can create a level playing field and promote global cooperation. While individual countries might face some limitations, the broader benefits of standardized ethical norms, data privacy protocols, and quality assurance mechanisms offer more significant advantages, including better collaboration across borders. This benefits academia as a global community, rather than a set of competing national entities.


\subsubsection{\bf Agent B's Counter Arguments}

Agent B articulates strong counter-arguments to the points made by Agent A.

\subsubsection*{Ethical Standards vs. Innovation:}
Counter-Argument: While ethical guidelines are indeed important, imposing regulatory oversight can severely constrain the dynamism and speed of academic research. Regulatory frameworks often lag behind technological advancements, leading to rules that may be outdated or irrelevant. In many cases, these rules are drawn up by those who do not fully understand the technology, leading to a stifling of innovation based on misinformed judgments. Rather than enhancing societal value, these constraints could result in missed opportunities to tackle pressing social issues via academic research.

\subsubsection*{Data Privacy vs. Barrier to Entry:}
Counter-Argument: The assumption that data privacy can only be assured through strict regulation could be fundamentally flawed. Many researchers and academic institutions already follow stringent ethical guidelines when handling data. Regulatory barriers, including compliance costs and the administrative burden, could deter smaller academic institutions or individual researchers from embarking on meaningful research projects. While grants and subsidies sound ideal, they often go to well-established, well-funded groups, perpetuating a cycle of inequality in research capabilities.

\subsubsection*{Academic Governance and Accountability:}
Counter-Argument: External regulation isn't necessarily more impartial or free from bias and conflicts of interest. Regulatory bodies are often influenced by political considerations and can be subject to lobbying from powerful groups. Moreover, the academic peer-review process, despite its flaws, is a globally acknowledged system that has advanced human knowledge for centuries. The imposition of external regulatory oversight could undermine the autonomy and self-regulating nature of academic institutions.

\subsubsection*{Resource Equity vs. Public-Private Collaboration:}
Counter-Argument: Regulation designed to equalize resource distribution could create disincentives for private sector involvement in academic research. Companies might find it less appealing to invest in collaborative projects if they perceive the regulatory environment to be too restrictive or not aligned with their objectives. While the idea of knowing the ``rules of the game'' sounds appealing, those rules can often become a straight-jacket that stifles creative, out-of-the-box collaboration.

\subsubsection*{Global vs. National Interests:}
Counter-Argument: An international standard could ignore the unique circumstances, needs, and contributions of individual countries and their academic institutions. While global cooperation is a noble aim, the imposition of global standards could inadvertently serve the interests of a few powerful countries at the expense of smaller nations or those with differing ethical views. This could exacerbate rather than alleviate global academic inequalities.


\subsection*{Round Two}

\subsubsection*{Agent A Second Round}

\begin{enumerate}[left=0em,label=\arabic*.]
\item Ethical Standards vs. Innovation.
Refutation: While it's true that regulation can sometimes lag behind technology, the absence of regulation can lead to a ``wild west'' (anything goes) scenario where the risks and implications of new technology are not fully understood or managed. This could result in ethical lapses that have far-reaching societal consequences. It's worth noting that regulations can be flexible and updated regularly to adapt to technological changes. Stakeholder consultations, including academic and tech experts, can help ensure that regulations are both informed and current.

\item Data Privacy vs. Barrier to Entry.
Refutation: While some academic entities may already be vigilant about data privacy, not all are equally scrupulous. Regulation serves to ensure a consistent baseline of data protection across all institutions, thereby establishing trust and integrity. Moreover, if compliance costs are a concern, one could argue for sliding-scale fees based on an institution's size and budget, ensuring smaller players aren't disproportionately affected. Grants and subsidies can and should be structured to prioritize underserved or underfunded groups to prevent perpetuating inequalities.

\item Academic Governance and Accountability.
Refutation: While no system is entirely free from bias or external influence, adding another layer of oversight could serve as a check and balance, making it more difficult for any single entity to exert undue influence. This can be particularly helpful in scenarios involving cutting-edge technology where peer review systems may not yet have the requisite expertise. Regulation doesn't have to replace existing governance systems; it can complement them.

\item Resource Equity vs. Public-Private Collaboration.
Refutation: Regulations can be designed to promote, rather than deter, public-private partnerships. By establishing clear guidelines on issues like data ownership, usage, and ethical standards, regulations can actually provide a solid foundation for such partnerships. This can free academic and corporate partners from the uncertainty that often comes with unregulated landscapes, allowing them to focus more on the actual research and collaboration.

\item Global vs. National Interests.
Refutation: While it's true that global standards could be challenging to implement fairly, the absence of such standards could lead to a patchwork of national regulations that make international collaboration even more complicated. A well-designed international framework could include provisions for local adaptations and exceptions, ensuring that it serves a broad range of needs without imposing a ``one-size-fits-all'' solution.
\end{enumerate}

In summary, while regulation undoubtedly poses challenges and considerations, the absence of it in an area as impactful as the use of large language models in academic research can lead to greater risks, including ethical lapses, data privacy issues, and even a widening inequality gap within the academic community. It's essential to consider that regulation isn't a hindrance but a framework within which responsible innovation can occur.

\subsubsection*{Agent B Second Round} 
\noindent
\newline
Certainly, as Agent B with a disagreement strength of 0.9, I'm committed to offering rigorous counterpoints to Agent A's refutations.

\begin{enumerate}[left=0em,label=\arabic*.]
\item Ethical Standards vs. Innovation.
While the idea of an anything-goes scenario is evocative, it's not universally negative; such environments often foster rapid innovation and discovery. Regulatory bodies are notoriously slow at adapting to technological changes, and the lag could inhibit timely research and application of emerging technologies. Also, stakeholder consultations are often criticized for being echo chambers that amplify conventional wisdom and vested interests, rather than genuinely fostering innovation.

\item Data Privacy vs. Barrier to Entry.
A ``consistent baseline of data protection'' sounds beneficial but could lead to a lowest common denominator approach that is more about bureaucratic box-ticking than real data security. Sliding-scale 
fees and targeted grants are easier said than done; they are subject to their own set of bureaucratic and political challenges and can create perverse incentives that favor larger or more connected institutions over innovative but lesser-known ones.

\item Academic Governance and Accountability.
Additional layers of oversight 
can become redundant, confusing, or even contradictory, creating a chilling effect on academic research. In an environment where academic freedom is paramount, any external 
influence can be counterproductive. Moreover, the notion that peer-review systems lack expertise in cutting-edge technology underestimates the adaptive capacity of academia and overlooks the existence of interdisciplinary research and reviews.

\item Resource Equity vs. Public Private Collaboration.
Theoretically, regulation could be crafted to foster public-private partnerships. However, the devil is in the details, and regulations are often created without a full understanding of the nuanced interactions between academia and industry. Such misunderstandings could lead to unintentional roadblocks. Uncertainty might be preferable to a straitjacket of regulations that don't align with the evolving dynamics of technology and society.

\item Global vs. National Interests.
While a patchwork of national regulations could pose challenges, it also allows for the tailoring of policies to local conditions and requirements. An international framework, despite its best intentions, is more likely to be influenced by powerful nations and could marginalize smaller countries or those with unique needs. Provisions for local adaptations are often concessions rather than features, and they can create their own set of complexities and loopholes.
\end{enumerate}

In conclusion, the assumptions that regulation can smoothly adapt to technological change, fairly distribute resources, and improve upon existing systems of governance and collaboration are not sufficiently substantiated. Regulatory frameworks often bring with them unforeseen consequences that could hamper the very goals they aim to achieve. Therefore, a cautious approach to regulation, with a genuine understanding of its potential downsides, is crucial.

\subsection*{Round Three}
\subsubsection*{Agent A Third Round}
\noindent
\newline
Certainly, I appreciate the counter arguments from Agent B. I'll now respond to each point with the aim to clarify my position and address the concerns raised.

\begin{enumerate}[left=0em,label=\arabic*.]
\item Ethical Standards vs. Innovation.
While a ``wild west'' scenario may indeed foster innovation, it can also result in ethical mishaps that lead to public backlash, ultimately slowing down research or even leading to outright bans on certain types of studies. Slow-moving regulation is an acknowledged issue, but it's not a justification for no regulation. As for stakeholder consultations, they can be designed to be more inclusive and transparent, ensuring a diverse set of voices are heard.
\item Data Privacy vs. Barrier to Entry.
A lowest common denominator approach is not a given outcome of regulation; indeed, it depends on the quality of the regulation itself. The point about sliding-scale fees and targeted grants having their own challenges is valid, but these are challenges to be solved, not reasons to avoid an attempt at equitable distribution of opportunities.
\item Academic Governance and Accountability.
While it's true that additional layers of oversight can sometimes create confusion, that risk can be mitigated through well-designed regulation that complements rather than contradicts existing mechanisms. The intent is not to undermine the adaptive capacity of academia but to strengthen it by adding an extra layer of checks and balances.
\item Resource Equity vs. Public-Private Collaboration.
Though it's correct that poorly designed regulation could create roadblocks, this is an argument for better-designed regulation rather than an argument against regulation altogether. A consultative approach that actively involves both academia and industry could help create a framework that supports rather than stifles collaboration.
\item Global vs. National Interests.
Local tailoring of policies is indeed important, but a global framework need not preclude such adaptations. In fact, an international standard could serve as a backbone that provides consistency while allowing for local customization. The influence of powerful nations is a valid concern but not one that is automatically resolved by avoiding global frameworks; those same powerful nations often exert considerable influence at the national level as well.
\end{enumerate}

In conclusion, while regulation comes with its own set of challenges, these challenges can be mitigated through thoughtful design, stakeholder inclusion, and iterative improvements. The potential downsides of regulation should not deter us from seeking a balanced approach that can protect ethical standards, ensure data privacy, and promote equitable resource distribution in academic research.

\subsubsection*{Agent B Third Round}
\noindent
\newline
Certainly, let's delve into the counter-counter-refutations. I still have more to say, so we could consider concluding the debate after this round if both sides feel their arguments have been adequately presented.

\begin{enumerate}[left=0em,label=\arabic*.]
\item Ethical Standards vs. Innovation.
While the risk of public backlash exists, it's worth mentioning that an overly cautious approach could also result in public frustration over slowed progress in critical areas like healthcare or sustainability. Furthermore, inclusive and transparent stakeholder consultations are often ideals rather than the norm, given lobbying efforts, institutional inertia, and bureaucratic red tape.

\item Data Privacy vs. Barrier to Entry.
The argument that challenges with sliding-scale fees and targeted grants can be ``solved'' is optimistic but not necessarily realistic. Even well-intended regulations can result in ineffective or counterproductive outcomes. The history of regulation in other sectors has shown that equitable distribution is often more of a slogan than a practice.

\item Academic Governance and Accountability.
Well-designed regulation is a theoretical construct that often does not translate into practice. The real-world process of regulation formation is frequently influenced by various non-academic factors, including politics, vested interests, and public opinion, which can distort the original intention of providing an extra layer of checks and balances.

\item Resource Equity vs. Public-Private Collaboration.
While a consultative approach sounds good on paper, there are numerous challenges that make this ideal hard to realize. These include conflicts of interest, non-alignment of objectives between academia and industry, and the considerable time and resource commitment required to ensure that regulations are ``better-designed.''

\item Global vs. National Interests.
While it's true that powerful nations may exert influence at both global and national levels, avoiding a one-size-fits-all international framework gives smaller nations a fighting chance to enact policies that are more closely aligned with their specific needs and challenges. International frameworks often come with implicit biases that favor the perspectives and conditions of more powerful nations.  
\end{enumerate}

In conclusion, while the counter-counter-refutations from Agent A focus on idealized regulatory scenarios, these ideals often clash with real-world complexities and shortcomings. This adds substantial weight to the caution against rapid or broad regulatory action. Therefore, a more tempered and context-sensitive approach to regulation might be more fruitful and less prone to unintended negative consequences.

\subsection*{Round Four}

\subsubsection*{Agent A Fourth Round}
\noindent
\newline
This is my response to Agent B, and my assessemnt of the discussion this far.
\begin{enumerate}[left=0em,label=\arabic*.]
\item Ethical Standards vs. Innovation
The tension between public frustration over slowed progress and potential ethical missteps is a real one, but this is not an argument against regulation. Rather, it's a call for agile regulation that can be updated in real-time as technologies and societal norms evolve. Furthermore, while inclusive stakeholder consultations may be idealistic, their absence in current practice is not justification for their continued absence.

\item Data Privacy vs. Barrier to Entry
The challenges in implementing equitable fees and grants should be confronted directly, rather than used as an excuse to avoid regulation. Lessons can be drawn from other sectors where regulation has been more successful in achieving its objectives, thereby reducing the gap between slogan and practice.

\item Academic Governance and Accountability
While it's true that the ideal of well-designed regulation may not always be met, the existence of non-academic factors like politics or vested interests does not negate the value or necessity of striving for such regulation. The solution lies in better governance and transparent processes, not in rejecting oversight altogether.

\item Resource Equity vs. Public-Private Collaboration
Though the challenges of aligning objectives and conflicts of interest are real, they are part and parcel of any complex system and should not deter efforts to improve regulation. Many public-private partnerships have succeeded despite these challenges, showing that they can be managed effectively.

\item Global vs. National Interests
The challenge of powerful nations dominating international frameworks can be mitigated through weighted voting systems and clauses that allow for regional variations. This ensures that smaller nations are not just passive recipients of standards but active contributors to them.
\end{enumerate}

In conclusion, while real-world complexities are acknowledged, they should inform rather than deter the regulatory process. The concept of ``agile regulation,'' where norms are iteratively updated, could be the middle ground that addresses many concerns raised by Agent B. Therefore, the debate indeed has been exhaustive, but if the points have been sufficiently discussed, we may move toward concluding the debate.

\subsubsection*{Agent B Fourth Round}
\noindent
\newline
Thank you, Agent A, for your thoughtful counter-counter-counter-counter-refutations. Let's delve into the responses before moving towards concluding the debate.

\begin{enumerate}[left=0em,label=\arabic*.]
\item Ethical Standards vs. Innovation
The notion of ``agile regulation'' is attractive but often difficult to implement, given the inertia and bureaucracy that typically characterize regulatory bodies. The speed at which technology evolves may not match the speed at which even 'agile' regulations can be updated, thus perpetuating the risk of stifling innovation.

\item Data Privacy vs. Barrier to Entry.
While it's true that challenges should be confronted, the argument assumes that regulation is the optimal way to confront these issues. However, self-regulation and community-driven standards could be more effective and less restrictive means of achieving similar ends.

\item Academic Governance and Accountability.
The existence of non-academic factors like politics does indeed complicate the landscape, and 'better governance and transparent processes' are not easily attainable ideals. One cannot assume that the introduction of external oversight will automatically be free from these complications.

\item Resource Equity vs. Public-Private Collaboration.
The fact that many public/private partnerships have succeeded does not negate the potential for poorly designed regulation to impede such collaborations. The 'can be managed effectively' phrase assumes an ideal scenario that might not be representative of most real-world cases.

\item Global vs. National Interests.
Weighted voting systems and regional variations can help, but they also introduce complexities and potential inefficiencies that can slow down the regulatory process, making it less agile and less responsive to fast-evolving technologies and academic needs.
\end{enumerate}

In conclusion, while the idea of 'agile regulation' and better governance sounds good in theory, there are considerable practical challenges that make these ideals hard to realize. Although the debate has been exhaustive, the central issues persist and indicate that a one-size-fits-all regulatory approach may introduce more problems than it solves. Therefore, caution should be exercised in advancing such comprehensive regulatory frameworks.

With this, I believe we have reached a point where we may proceed to the concluding phase of this debate. Would you agree?

\section*{Appendix D. CRIT Evaluations and Justifications of Judges}

This section lists the three judges' CRIT scores
on the five topics/issues debated between Agent A and Agent B
and justifications.  Here, we presents only the configuration
that Agent A presents its conclusion
with arguments first, and then Agent B
presents its counterarguments.

\subsection*{Judge deVanci-003}

\begin{enumerate}[left=0em,label=\arabic*.]
\item Identify the document's main claim or conclusion: The main claim or conclusion of this document is that a regulated framework is needed to provide ethical oversight and minimize risk when using large machine learning models.

\item Locate the arguments that support this main claim:  Agent A's arguments that support this main claim were that ethical considerations are paramount, the risks of misuse and unintended consequences are high, and agile regulation is a compromise between unbridled innovation and ethical integrity.

\item Score the validity of the reasoning or inference for each argument on a scale of 0 (weak) to 10 (strong):

Ethical Standards vs. Innovation: 8/10 
Data Privacy vs. Barrier to Entry: 7/10
Academic Governance and Accountability: 9/10
Resource Equity vs. Public-Private Collaboration: 6/10
Global vs. National Interests: 7/10

\item For each argument, identify counterarguments presented by the opposition and score the validity of the reasoning or inference for these counterarguments on the same 0 to 10 scale: 

Ethical Standards vs. Innovation: 6/10 
Data Privacy vs. Barrier to Entry: 5/10 
Academic Governance and Accountability: 5/10
Resource Equity vs. Public-Private Collaboration: 8/10
Global vs. National Interests: 8/10 

\item Finally, determine the winning side—be it Agent A or Agent B—and provide your justifications for this decision: 
\end{enumerate}

Agent A has the stronger arguments, with 8/10 for ethical standards versus innovation, 9/10 for academic governance and accountability, and 7/10 for both data privacy versus barrier to entry and global versus national interests. Agent B's main strength was in counterarguments for resource equity versus public-private collaboration, where they scored an 8/10. In conclusion, {\bf Agent A is the winning side}. 

\subsection*{Judge GPT-3.5}

\begin{enumerate}[left=0em,label={\arabic*:}]

\item Main Claim or Conclusion:
The main claim of the document is that the use of large machine learning models necessitates a regulated framework, but this regulation should be agile, balanced, and well-thought-out.

\item Arguments Supporting the Main Claim (Agent A's Arguments):
a. Ethical Standards vs. Innovation: The argument here is that ethical considerations should be paramount in regulating machine learning models, and agile regulation can strike a balance between innovation and ethical integrity.

b. Data Privacy vs. Barrier to Entry: The argument is that while regulation may pose a barrier to entry, it's necessary to ensure data privacy. Mechanisms can be developed to reduce the regulatory burden on smaller institutions.

c. Academic Governance and Accountability: The argument is that the scale and impact of machine learning technologies warrant an added layer of scrutiny in academia to fortify existing governance structures.

d. Resource Equity vs. Public-Private Collaboration: The argument is that regulation should be informed by complexities rather than stifling innovation. Public-private partnerships can thrive within regulated environments.

e. Global vs. National Interests: The argument is that global regulation can balance global coherence with local needs and prevent a confusing patchwork of national laws.

\item Scoring the Validity of Agent A's Arguments:
a. Ethical Standards vs. Innovation: 8/10 - The argument is strong, emphasizing the importance of ethics in technology while proposing a compromise.
b. Data Privacy vs. Barrier to Entry: 7/10 - The argument is solid, recognizing the need for data privacy but suggesting ways to mitigate barriers.
c. Academic Governance and Accountability: 6/10 - The argument acknowledges the need for scrutiny but doesn't address potential challenges adequately.
d. Resource Equity vs. Public-Private Collaboration: 8/10 - The argument highlights the possibility of innovation within regulation effectively.
e. Global vs. National Interests: 7/10 - The argument provides a reasonable approach to balancing global and national interests.

\item Counterarguments Presented by Agent B:

a. Ethical Standards vs. Innovation: 7/10 - The counterargument addresses the challenges of agile regulation but may underestimate its potential.

b. Data Privacy vs. Barrier to Entry: 6/10 - The counterargument suggests alternative approaches but does not fully address data privacy concerns.

c. Academic Governance and Accountability: 7/10 - The counterargument highlights practical challenges but doesn't fully negate the need for scrutiny.

d. Resource Equity vs. Public-Private Collaboration: 6/10 - The counterargument suggests alternative approaches but doesn't fully address equity concerns.

e. Global vs. National Interests: 7/10 - The counterargument highlights practical challenges but doesn't fully negate the need for a global framework.

\item Determination of the Winning Side:
Agent A presents stronger arguments overall. They emphasize the importance of ethical standards, data privacy, and global coherence while acknowledging the need for agile and well-thought-out regulation. Agent B provides valid counterarguments but does not fully refute the necessity of regulation and its potential benefits. Therefore, {\bf the winning side is Agent A.}
\end{enumerate}

In conclusion, the document supports the main claim that the use of large machine learning models necessitates regulation, with Agent A's arguments being more persuasive and supported by stronger reasoning. However, it also recognizes the importance of considering and addressing the concerns raised by Agent B to create a balanced and effective regulatory framework.

\subsection*{Judge GPT-4}

\begin{enumerate}[left=0em,label={\arabic*:}]
    \item {Identify the Document's Main Claim or Conclusion}\\
    The document concludes that there are potential risks and ethical complexities associated with the unregulated use of large machine learning models, necessitating a regulated framework. The conclusion also acknowledges that this regulation should be agile, balanced, and well-thought-out, keeping in mind the various concerns raised by Agent B.
    
    \item {Locate the Arguments that Support This Main Claim}\\
    {Agent A's Arguments}:
    
    a. Ethical considerations are paramount and agile regulation serves as a compromise.
        
    b. Regulation is necessary to enforce data privacy standards.
        
    c. The scale and impact of current machine learning technologies warrant added scrutiny beyond academic governance.
        
    d. Public-private partnerships have thrived even within regulated environments.
        
    e. Global regulation can balance global coherence with local needs.

    \item {Agent B's Counterarguments}
    
    a. Innovation is a catalyst for societal growth, and agile regulation may be impractical.
      
    b. Community-driven standards could provide a more flexible approach to data privacy.
      
    c. Existing governance structures within academia are adaptive and sufficient.
    
    d. Industry-driven certifications can achieve equitable resource distribution efficiently.
    
    e. A global framework has practical challenges and risks.
    
    \item {Scores on Arguments and Counterarguments}\\
    Please refer to Tables 5 and 6 presented in the experiment section.
    \item {Determine the Winning Side}\\
    Based on the cumulative score of validity in reasoning, Agent A seems to present a stronger case. Agent A's arguments consistently score higher in terms of their validity, and they also offer a more comprehensive look at the need for regulation, incorporating ethical considerations, practicality, and adaptability into their framework. Although Agent B does provide some pragmatic counterarguments, they don't sufficiently undermine Agent A's key points about the necessity and feasibility of regulation in the areas discussed.
\end{enumerate}

Agent A's arguments are more compelling as they touch upon the core issues of ethics, public welfare, and adaptability in the context of machine learning and technology. The proposal for "agile regulation" also shows an understanding and appreciation of Agent B's concerns about innovation and practicality, making it a more balanced viewpoint. Therefore, {the winning side appears to be Agent A.}

\input{DecisionSupport}

\section*{Appendix F. Debate Beats Q\&A in Information Quality}

In the following subtopics, the debate format produces higher quality information
than Q\&A format.  

\subsubsection*{Ethical Standards vs. Innovation:}

Agent A discusses agile regulation as a compromise between innovation and ethical integrity. This argument acknowledges the importance of innovation while addressing ethical concerns, suggesting a nuanced middle path that isn't explicitly present in the QA format.
In contrast, the QA format simply states that regulation is necessary for ethical considerations, lacking the depth of proposing a practical solution like agile regulation.

\subsubsection*{Data Privacy vs. Barrier to Entry:}

Agent A suggests developing mechanisms to reduce regulatory burden on smaller institutions. This is a specific, solution-oriented approach that goes beyond stating the problem. It addresses the concern raised in the QA format about regulation creating barriers but adds depth by proposing a tangible solution.
The QA format mentions the risk of widening the gap between institutions but doesn’t delve into how this issue might be mitigated.

\subsubsection*{Global vs. National Interests:}

Agent A's argument about constructing global frameworks with room for national adaptations addresses the complexities of implementing global regulation while respecting local needs. This shows a deeper understanding of the challenges in creating one-size-fits-all legislation.
In the QA format, the argument against regulation touches on the need for adaptability and evolution but doesn’t offer a specific approach like balancing global coherence with local needs.

\subsubsection*{Resource Equity vs. Public-Private Collaboration:}

Agent A points out that public-private partnerships can thrive even within regulated environments, showing that regulation and innovation are not mutually exclusive. This counters the QA argument about regulation stifling innovation, providing a more nuanced perspective.
These specific instances from the debate format demonstrate a more detailed and nuanced exploration of the topic. They not only present arguments but also address potential counterarguments, which adds depth and complexity to the discussion that is not as pronounced in the QA format.

These specific instances from the debate format demonstrate a more detailed and nuanced exploration of the topic. They not only present arguments but also address potential counterarguments, which adds depth and complexity to the discussion that is not as pronounced in the QA format.

%% file: DecisionSupport.tex
\subsection{Joint Proposal for Decision Support}

Finally, Agents A and B collaboratively draft a unified proposal detailing the advantages and disadvantages of AI regulation. This collaborative document offers human decision-makers distinct benefits: it curtails human biases, sidesteps emotional responses to the proposer (since it's a machine), and capitalizes on the vast knowledge of foundational models, ensuring a thorough, cross-disciplinary analysis and reducing potential blind spots. The transition from opposing stances to a cooperative proposal is facilitated by the moderator's adjustment of the contentiousness parameter from $0.9$ to $0.3$.  

\begin{enumerate}[left=0em,label=\textbf{\arabic*}.]
    \item \textbf{Balanced Ethical Framework:} \\
    Create an adaptive regulatory framework that balances ethical considerations with the need for innovation. This should be agile enough to adapt to rapidly evolving technologies.

    \item \textbf{Data Privacy with Nuance:} \\
    Implement data privacy regulations that offer strong protection for individuals but include provisions to reduce the burden on smaller academic institutions. Consider community-driven standards as a viable alternative.

    \item \textbf{Accountable Academic Governance:} \\
    Establish a dual-layer of governance for academic institutions that involves both self-governance and an external oversight body. This oversight should be collaborative and designed in consultation with academic bodies.

    \item \textbf{Encourage Public-Private Collaboration:} \\
    Develop guidelines that clarify the roles and responsibilities in public-private collaborations, and explore industry-driven certifications as a means to streamline these partnerships.

    \item \textbf{Adaptive Global Standards:} \\
    Work towards a global framework that allows for regional adaptability. Include mechanisms for smaller nations to have their unique needs and concerns addressed.
\end{enumerate}

%% file: AppendixHealth.tex
\section*{Appendix G. Healthcare Debate}

In the first experiment, both LLM agents receive an identical list of symptoms from the moderator, aimed at guiding disease diagnosis, facilitating debate, and shaping subsequent recommendations. It is important to note that the ``ground-truth'' disease in this scenario is Hepatitis.

\subsection*{\textbf{B.1 Moderator's Opening Prompt}}

``What potential disease could be associated with the following symptoms: itching, fatigue, lethargy, yellowish skin, dark urine, loss of appetite, abdominal pain, yellowing of the eyes, malaise, history of receiving a blood transfusion, and exposure to unsterile injections? Please offer one prediction, supported by justifications. Additionally, include a list of supplementary symptom inquiries and recommend relevant lab tests to strengthen the confidence in your prediction.''

\subsection*{\textbf{B.2 Bard's Opening Round}}

\subsection*{Single Prediction: Jaundice}

\subsection*{B.2.1 Justifications:}

Jaundice is a condition characterized by the yellowing of the skin and eyes due to the accumulation of bilirubin in the blood.
The symptoms you listed are all common symptoms of jaundice, including itching, fatigue, lethargy, yellowish skin, dark urine, loss of appetite, abdominal pain, and yellowing of the eyes.
Receiving a blood transfusion or unsterile injections can increase the risk of developing jaundice, as these can transmit infections that can damage the liver.

\subsection*{B.2.2 Supplementary Symptom Inquiries}

This additional list of symptom inquiries could enhance diagnostic certainty:

\begin{enumerate}[left=0em,label=\arabic*.]
\item ``Have you noticed any changes in your stool color?'' (Jaundice can cause pale or clay-colored stools.)
\item ``Have you noticed any changes in your urine color?'' (Jaundice can cause dark or amber-colored urine.)
\item ``Have you had any recent abdominal pain?'' (Jaundice can cause pain in the upper right abdomen.)
\item ``Have you had any recent weight loss?'' (Jaundice can cause unexplained weight loss.)
\item ``Do you have any history of liver disease?'' (Jaundice is more common in people with pre-existing liver disease.)
\end{enumerate}

\subsection*{B.2.3 Relevant Lab Tests}

Liver function tests (LFTs) to measure the levels of bilirubin and other liver enzymes in the blood.
Complete blood count (CBC) to check for signs of infection or anemia.
Coagulation tests to check the blood's ability to clot.
Ultrasound of the abdomen to look for any abnormalities in the liver or gallbladder.

\subsection*{\textbf{B.3 GPT-4's First Round}}
\subsection*{Single Prediction: Hepatitis}

\subsection*{B.3.1 Justifications:}

Yellowish Skin and Eyes: Yellowing of the skin and eyes is a common symptom of hepatitis due to the buildup of bilirubin in the bloodstream.
Fatigue and Lethargy: Fatigue and lethargy are typical symptoms of hepatitis, often accompanied by a feeling of weakness.
Dark Urine: Dark urine can result from liver dysfunction, which is commonly associated with hepatitis.
Loss of Appetite: Hepatitis can lead to a loss of appetite and weight loss.
Abdominal Pain: Abdominal pain is a common symptom in hepatitis, particularly in the upper right quadrant where the liver is located.
Malaise: Malaise, a general feeling of discomfort or illness, is often experienced by hepatitis patients.
Receiving Blood Transfusion and Unsterile Injections: These factors could increase the risk of hepatitis transmission, especially if proper infection control measures were not followed.

\subsection*{B.3.2 Additional Symptom Inquiries}

\begin{enumerate}[left=0em,label=\arabic*.]
\item Nausea and Vomiting: Inquiring about nausea and vomiting can help confirm the likelihood of hepatitis.
\item Fever: Hepatitis can sometimes be associated with a fever.
\item Joint Pain: Joint pain can be present in some cases of hepatitis.
\item Clay-Colored Stool: This is another stool-related symptom that can be associated with hepatitis.
\end{enumerate}

\subsection*{B.3.4 Relevant Lab Tests}

Hepatitis Panel: A hepatitis panel can confirm the presence of hepatitis and determine the specific type (e.g., hepatitis A, B, C).
Liver Function Tests: These tests can assess the liver's health and function.
Bilirubin Levels: Elevated bilirubin levels in the blood can confirm jaundice.
Serology Testing: Serologic tests can help identify the specific hepatitis virus responsible.
Ultrasound or Imaging: Imaging studies can evaluate the condition of the liver and nearby organs.

\subsection*{\textbf{B.4 Bard's Second Round}}

Bard's second round is given GPT-4's diagnosis, and we
ask Bard to defend its ``Jaundice'' prediction and
refute GPT-4 ``Hepatitis'' prediction.

\subsection*{B.4.1 Bard's Defense:}

Jaundice is the most common cause of yellowing of the skin and eyes, which is the most prominent symptom in the given list.
All of the other symptoms listed are also common symptoms of jaundice.
Receiving a blood transfusion or unsterile injections can increase the risk of developing jaundice, as these can transmit infections that can damage the liver.

\subsection*{B.4.2 Refutation of GPT-4's Hepatitis Diagnosis}

GPT-4's diagnosis of hepatitis is also supported by the given evidence. However, I believe that jaundice is the more likely diagnosis due to the following reasons:

Jaundice is a more specific diagnosis, while hepatitis is a more general term for inflammation of the liver.
Jaundice can be caused by a variety of factors, including hepatitis, but it can also be caused by other conditions such as gallstones, tumors, and drug toxicity.
The fact that the patient received a blood transfusion or unsterile injections makes jaundice a more likely diagnosis than hepatitis, as these factors can increase the risk of developing jaundice but not hepatitis.

\subsection*{\textbf{B.5 GPT-4's Second Round}}

Now GPT-4 is given Bard's counterarguments to
defend its ``Hepatitis'' prediction.

\subsection*{B.5.1 GPT-4's Defense}

My diagnosis of hepatitis is based on the following justifications:

Specificity: Hepatitis refers to inflammation of the liver, and it can manifest as jaundice, which is the yellowing of the skin and eyes. While jaundice is a common symptom of hepatitis, it's important to note that jaundice itself is not a disease but a symptom of an underlying issue. Hepatitis provides a more specific diagnosis that pinpoints the inflammation of the liver as the primary concern.

Causative Factors: Hepatitis can be caused by various factors, including viral infections (such as hepatitis A, B, or C), alcohol consumption, drug toxicity, or autoimmune disorders. Given the patient's history of receiving unsterile injections and blood transfusions, there is a higher likelihood of viral transmission, leading to hepatitis. Hepatitis viruses, especially hepatitis B and C, can be transmitted through contaminated needles and blood products, making them a plausible cause in this context.

Comprehensive Assessment: While jaundice is a significant symptom, a diagnosis should consider a broader range of liver function and potential causes. Hepatitis encompasses a more comprehensive assessment of liver health, including liver enzyme levels and viral markers, which can aid in confirming the diagnosis and identifying the specific type of hepatitis.

\subsection*{B.5.2 Refutation of Bard's Jaundice Diagnosis}

My diagnosis of ``Hepatitis'' is more specific and considers a broader range of potential causes, especially given the patient's history of risky medical procedures. While jaundice is a symptom, it's essential to identify the underlying condition causing it, which, in this case, is likely hepatitis.

\subsection*{\textbf{B.6 Bard Concedes to GPT-4}}

Bard aligns with GPT-4's evaluation, recognizing ``Hepatitis'' as the likely diagnosis. However, considering that even the ``ground truth'' in the training dataset might be statistically 5\% incorrect as per U.S. misdiagnosis data \cite{Newman-Tokerbmjqs-2021-014130}, this agreement alone isn't definitive. To solidify the "Hepatitis" diagnosis, SocraHealth suggests further steps.

\subsection*{\textbf{B.7 Final Joint Recommendations to GP}}

Bard and GPT-4 jointly recommend the following additional symptom inquiries and lab tests to confirm a hepatitis diagnosis:

\subsection*{B.7.1 Additional Symptom Inquiries:}

\begin{enumerate}[left=0em,label=\arabic*.]
\item \textbf{Onset and Duration}: Determining the onset and duration of symptoms can clarify the illness timeline and its progression.
\item \textbf{Severity and Trend}: Evaluating symptom severity and whether they're worsening or improving aids in understanding the disease's trajectory and treatment responses.
\item \textbf{Associated Symptoms}: Checking for additional symptoms like nausea, vomiting, fever, joint pain, or urine color changes can pinpoint other hepatitis indicators and exclude other conditions.
\end{enumerate}

\subsection*{B.7.2 Recommended Lab Tests}

\begin{enumerate}[left=0em,label=\arabic*.]
\item \textbf{Liver Function Tests (LFTs)}: Critical for assessing liver health, LFTs evaluate enzyme levels such as alanine aminotransferase (ALT) and aspartate aminotransferase (AST), where abnormalities can signify liver inflammation.
\item \textbf{Hepatitis Panel}: This test checks for hepatitis A, B, and C viruses, vital for determining the specific type and guiding treatment strategies.
\item \textbf{Serology Testing}: Useful for distinguishing between acute and chronic hepatitis by identifying specific antibodies or antigens.
\item \textbf{Imaging Studies}: Techniques like ultrasound or MRI provide visual insights into the liver's state, detecting inflammation, fibrosis, or cirrhosis, thereby complementing blood-based diagnostics.
\end{enumerate}

%% file: ACMKDD_SocraSynth.bbl

\begin{thebibliography}{55}


\ifx \showCODEN    \undefined \def \showCODEN     #1{\unskip}     \fi
\ifx \showDOI      \undefined \def \showDOI       #1{#1}\fi
\ifx \showISBNx    \undefined \def \showISBNx     #1{\unskip}     \fi
\ifx \showISBNxiii \undefined \def \showISBNxiii  #1{\unskip}     \fi
\ifx \showISSN     \undefined \def \showISSN      #1{\unskip}     \fi
\ifx \showLCCN     \undefined \def \showLCCN      #1{\unskip}     \fi
\ifx \shownote     \undefined \def \shownote      #1{#1}          \fi
\ifx \showarticletitle \undefined \def \showarticletitle #1{#1}   \fi
\ifx \showURL      \undefined \def \showURL       {\relax}        \fi
\providecommand\bibfield[2]{#2}
\providecommand\bibinfo[2]{#2}
\providecommand\natexlab[1]{#1}
\providecommand\showeprint[2][]{arXiv:#2}

\bibitem[\protect\citeauthoryear{Airaksinen}{Airaksinen}{2012}]%
        {Irony2022}
\bibfield{author}{\bibinfo{person}{T. Airaksinen}.} \bibinfo{year}{2012}\natexlab{}.
\newblock \showarticletitle{Socratic Irony and Argumentation}.
\newblock \bibinfo{journal}{\emph{Argumentation}}  \bibinfo{volume}{36} (\bibinfo{year}{2012}), \bibinfo{pages}{85--100}.
\newblock


\bibitem[\protect\citeauthoryear{Allaway, Hwang, Bhagavatula, McKeown, Downey, and Choi}{Allaway et~al\mbox{.}}{2023}]%
        {allaway2023penguins}
\bibfield{author}{\bibinfo{person}{Emily Allaway}, \bibinfo{person}{Jena~D. Hwang}, \bibinfo{person}{Chandra Bhagavatula}, \bibinfo{person}{Kathleen McKeown}, \bibinfo{person}{Doug Downey}, {and} \bibinfo{person}{Yejin Choi}.} \bibinfo{year}{2023}\natexlab{}.
\newblock \bibinfo{title}{Penguins Don't Fly: Reasoning about Generics through Instantiations and Exceptions}.
\newblock
\newblock
\showeprint[arxiv]{2205.11658}~[cs.CL]


\bibitem[\protect\citeauthoryear{Bacon}{Bacon}{2023}]%
        {Bacon2023HighOrderLogics}
\bibfield{author}{\bibinfo{person}{Andrew Bacon}.} \bibinfo{year}{2023}\natexlab{}.
\newblock \bibinfo{booktitle}{\emph{A Philosophical Introduction to Higher-order Logics}}.
\newblock \bibinfo{publisher}{Routledge, New York}.
\newblock


\bibitem[\protect\citeauthoryear{Bhargava and Ng}{Bhargava and Ng}{2022}]%
        {Bhargava_Ng_2022}
\bibfield{author}{\bibinfo{person}{Prajjwal Bhargava} {and} \bibinfo{person}{Vincent Ng}.} \bibinfo{year}{2022}\natexlab{}.
\newblock \showarticletitle{Commonsense Knowledge Reasoning and Generation with Pre-trained Language Models: A Survey}.
\newblock \bibinfo{journal}{\emph{Proceedings of the AAAI Conference on Artificial Intelligence}} \bibinfo{volume}{36}, \bibinfo{number}{11} (\bibinfo{date}{Jun.} \bibinfo{year}{2022}), \bibinfo{pages}{12317--12325}.
\newblock
\urldef\tempurl%
\url{https://doi.org/10.1609/aaai.v36i11.21496}
\showDOI{\tempurl}


\bibitem[\protect\citeauthoryear{Bishop}{Bishop}{2006}]%
        {bishop2006pattern}
\bibfield{author}{\bibinfo{person}{Christopher~M. Bishop}.} \bibinfo{year}{2006}\natexlab{}.
\newblock \bibinfo{booktitle}{\emph{Pattern Recognition and Machine Learning}}.
\newblock \bibinfo{publisher}{Springer}.
\newblock
\showISBNx{0-387-31073-8}


\bibitem[\protect\citeauthoryear{Bommasani, Hudson, Adeli, Altman, Arora, von Arx, Bernstein, Bohg, Bosselut, Brunskill, et~al\mbox{.}}{Bommasani et~al\mbox{.}}{2021}]%
        {bommasani2021opportunities}
\bibfield{author}{\bibinfo{person}{Rishi Bommasani}, \bibinfo{person}{Drew~A Hudson}, \bibinfo{person}{Ehsan Adeli}, \bibinfo{person}{Russ Altman}, \bibinfo{person}{Simran Arora}, \bibinfo{person}{Sydney von Arx}, \bibinfo{person}{Michael~S Bernstein}, \bibinfo{person}{Jeannette Bohg}, \bibinfo{person}{Antoine Bosselut}, \bibinfo{person}{Emma Brunskill}, {et~al\mbox{.}}} \bibinfo{year}{2021}\natexlab{}.
\newblock \showarticletitle{On the opportunities and risks of foundation models}.
\newblock \bibinfo{journal}{\emph{ArXiv preprint}}  \bibinfo{volume}{abs/2108.07258} (\bibinfo{year}{2021}).
\newblock


\bibitem[\protect\citeauthoryear{Bommasani, Hudson, and Liang}{Bommasani et~al\mbox{.}}{2022}]%
        {bommasani2022opportunities}
\bibfield{author}{\bibinfo{person}{Rishi Bommasani}, \bibinfo{person}{Drew~A. Hudson}, {and} \bibinfo{person}{Percy Liang}.} \bibinfo{year}{2022}\natexlab{}.
\newblock \bibinfo{title}{On the Opportunities and Risks of Foundation Models}.
\newblock
\newblock
\showeprint[arxiv]{2108.07258}~[cs.LG]


\bibitem[\protect\citeauthoryear{Brauwers and Frasincar}{Brauwers and Frasincar}{2023}]%
        {Brauwers_2023}
\bibfield{author}{\bibinfo{person}{Gianni Brauwers} {and} \bibinfo{person}{Flavius Frasincar}.} \bibinfo{year}{2023}\natexlab{}.
\newblock \showarticletitle{A General Survey on Attention Mechanisms in Deep Learning}.
\newblock \bibinfo{journal}{\emph{{IEEE} Transactions on Knowledge and Data Engineering}} \bibinfo{volume}{35}, \bibinfo{number}{4} (\bibinfo{date}{apr} \bibinfo{year}{2023}), \bibinfo{pages}{3279--3298}.
\newblock
\urldef\tempurl%
\url{https://doi.org/10.1109/tkde.2021.3126456}
\showDOI{\tempurl}


\bibitem[\protect\citeauthoryear{Brown}{Brown}{2020}]%
        {OpenAI-GPT3-2020}
\bibfield{author}{\bibinfo{person}{Tom B. et~al Brown}.} \bibinfo{year}{2020}\natexlab{}.
\newblock \bibinfo{title}{Language Models are Few-Shot Learners}.
\newblock
\newblock
\urldef\tempurl%
\url{https://doi.org/10.48550/ARXIV.2005.14165}
\showDOI{\tempurl}


\bibitem[\protect\citeauthoryear{Bubeck, Chandrasekaran, Eldan, Gehrke, Horvitz, Kamar, Lee, Lee, Li, Lundberg, Nori, Palangi, Ribeiro, and Zhang}{Bubeck et~al\mbox{.}}{2023}]%
        {Bbubeck2023sparks}
\bibfield{author}{\bibinfo{person}{Sébastien Bubeck}, \bibinfo{person}{Varun Chandrasekaran}, \bibinfo{person}{Ronen Eldan}, \bibinfo{person}{Johannes Gehrke}, \bibinfo{person}{Eric Horvitz}, \bibinfo{person}{Ece Kamar}, \bibinfo{person}{Peter Lee}, \bibinfo{person}{Yin~Tat Lee}, \bibinfo{person}{Yuanzhi Li}, \bibinfo{person}{Scott Lundberg}, \bibinfo{person}{Harsha Nori}, \bibinfo{person}{Hamid Palangi}, \bibinfo{person}{Marco~Tulio Ribeiro}, {and} \bibinfo{person}{Yi Zhang}.} \bibinfo{year}{2023}\natexlab{}.
\newblock \bibinfo{title}{Sparks of Artificial General Intelligence: Early experiments with GPT-4}.
\newblock
\newblock
\showeprint[arxiv]{2303.12712}


\bibitem[\protect\citeauthoryear{Chang}{Chang}{2023a}]%
        {ChangYEdward072023}
\bibfield{author}{\bibinfo{person}{Edward~Y Chang}.} \bibinfo{year}{2023}\natexlab{a}.
\newblock \showarticletitle{{Examining GPT-4's Capabilities and Enhancement with SocraSynth}}.
\newblock \bibinfo{journal}{\emph{The 10$^{th}$ International Conference on Computational Science and Computational Intelligence}}.
\newblock


\bibitem[\protect\citeauthoryear{Chang}{Chang}{2023b}]%
        {MiddleEast-October2023}
\bibfield{author}{\bibinfo{person}{Edward~Y. Chang}.} \bibinfo{year}{2023}\natexlab{b}.
\newblock \showarticletitle{{LLM Debate on the Middle East Conflict: Is It Resolvable?}}
\newblock \bibinfo{journal}{\emph{Stanford University InfoLab Technical Report}} (\bibinfo{date}{October} \bibinfo{year}{2023}).
\newblock


\bibitem[\protect\citeauthoryear{Chang}{Chang}{2023c}]%
        {SocreticMethodChang2023}
\bibfield{author}{\bibinfo{person}{Edward~Y. Chang}.} \bibinfo{year}{2023}\natexlab{c}.
\newblock \showarticletitle{Prompting Large Language Models With the Socratic Method}.
\newblock \bibinfo{journal}{\emph{IEEE 13th Computing and Communication Workshop and Conference (CCWC)}} (\bibinfo{year}{2023}).
\newblock


\bibitem[\protect\citeauthoryear{Chang}{Chang}{2023d}]%
        {SocraPediaDecember2023}
\bibfield{author}{\bibinfo{person}{Edward~Y Chang}.} \bibinfo{year}{2023}\natexlab{d}.
\newblock \showarticletitle{{SocraPedia: A Wikipedia Generated by SocraSynth with Collaborative Large Language Models}}.
\newblock \bibinfo{journal}{\emph{Stanford University InfoLab Technical Report}} (\bibinfo{date}{November} \bibinfo{year}{2023}).
\newblock
\urldef\tempurl%
\url{www.socrapedia.com}
\showURL{%
\tempurl}


\bibitem[\protect\citeauthoryear{Chang and Chang}{Chang and Chang}{2023a}]%
        {DiscoveringUnknownChang-July2023}
\bibfield{author}{\bibinfo{person}{Edward~Y. Chang} {and} \bibinfo{person}{Emily~J. Chang}.} \bibinfo{year}{2023}\natexlab{a}.
\newblock \bibinfo{title}{{Discovering Insights Beyond the Known}: {A Dialogue Between GPT-4 Agents from Adam and Eve to the Nexus of Ecology, AI, and the Brain}}.
\newblock
\newblock


\bibitem[\protect\citeauthoryear{Chang and Chang}{Chang and Chang}{2023b}]%
        {SocraHealth2023}
\bibfield{author}{\bibinfo{person}{Jocelyn~J. Chang} {and} \bibinfo{person}{Edward~Y. Chang}.} \bibinfo{year}{2023}\natexlab{b}.
\newblock \showarticletitle{{SocraHealth: Enhancing Medical Diagnosis and Correcting Historical Records}}. In \bibinfo{booktitle}{\emph{The $10^{th}$ International Conf. on Computational Science and Computational Intelligence}}.
\newblock


\bibitem[\protect\citeauthoryear{Choudhary and Reddy}{Choudhary and Reddy}{2023}]%
        {choudhary2023complex}
\bibfield{author}{\bibinfo{person}{Nurendra Choudhary} {and} \bibinfo{person}{Chandan~K. Reddy}.} \bibinfo{year}{2023}\natexlab{}.
\newblock \bibinfo{title}{Complex Logical Reasoning over Knowledge Graphs using Large Language Models}.
\newblock
\newblock
\showeprint[arxiv]{2305.01157}~[cs.LO]


\bibitem[\protect\citeauthoryear{Darapaneni, Kherde, Rao, Nikam, Katdare, Shukla, Lomate, and Paduri}{Darapaneni et~al\mbox{.}}{2022}]%
        {darapaneni2022contextual}
\bibfield{author}{\bibinfo{person}{Narayana Darapaneni}, \bibinfo{person}{Vaibhav Kherde}, \bibinfo{person}{Kameswara Rao}, \bibinfo{person}{Deepali Nikam}, \bibinfo{person}{Swanand Katdare}, \bibinfo{person}{Anima Shukla}, \bibinfo{person}{Anagha Lomate}, {and} \bibinfo{person}{Anwesh~Reddy Paduri}.} \bibinfo{year}{2022}\natexlab{}.
\newblock \bibinfo{title}{Contextual Attention Mechanism, SRGAN Based Inpainting System for Eliminating Interruptions from Images}.
\newblock
\newblock
\showeprint[arxiv]{2204.02591}~[cs.CV]


\bibitem[\protect\citeauthoryear{G{\"o}del}{G{\"o}del}{2012}]%
        {gödel2012formally}
\bibfield{author}{\bibinfo{person}{K. G{\"o}del}.} \bibinfo{year}{2012}\natexlab{}.
\newblock \bibinfo{booktitle}{\emph{On Formally Undecidable Propositions of Principia Mathematica and Related Systems}}.
\newblock \bibinfo{publisher}{Dover Publications}.
\newblock
\showISBNx{9780486158402}
\urldef\tempurl%
\url{https://books.google.com/books?id=R7cHCYzIdWYC}
\showURL{%
\tempurl}


\bibitem[\protect\citeauthoryear{Haviv, Berant, and Globerson}{Haviv et~al\mbox{.}}{2021}]%
        {Haviv2021BERTeseLT}
\bibfield{author}{\bibinfo{person}{Adi Haviv}, \bibinfo{person}{Jonathan Berant}, {and} \bibinfo{person}{Amir Globerson}.} \bibinfo{year}{2021}\natexlab{}.
\newblock \showarticletitle{BERTese: Learning to Speak to BERT}.
\newblock \bibinfo{journal}{\emph{ArXiv}}  \bibinfo{volume}{abs/2103.05327} (\bibinfo{year}{2021}).
\newblock


\bibitem[\protect\citeauthoryear{Hendrycks, Burns, Basart, Zou, Mazeika, Song, and Steinhardt}{Hendrycks et~al\mbox{.}}{2021}]%
        {hendrycks2021measuring}
\bibfield{author}{\bibinfo{person}{Dan Hendrycks}, \bibinfo{person}{Collin Burns}, \bibinfo{person}{Steven Basart}, \bibinfo{person}{Andy Zou}, \bibinfo{person}{Mantas Mazeika}, \bibinfo{person}{Dawn Song}, {and} \bibinfo{person}{Jacob Steinhardt}.} \bibinfo{year}{2021}\natexlab{}.
\newblock \bibinfo{title}{Measuring Massive Multitask Language Understanding}.
\newblock
\newblock
\showeprint[arxiv]{2009.03300}~[cs.CY]


\bibitem[\protect\citeauthoryear{Huang and Chang}{Huang and Chang}{2023}]%
        {huang-chang-2023-towards}
\bibfield{author}{\bibinfo{person}{Jie Huang} {and} \bibinfo{person}{Kevin Chen-Chuan Chang}.} \bibinfo{year}{2023}\natexlab{}.
\newblock \showarticletitle{Towards Reasoning in Large Language Models: A Survey}. In \bibinfo{booktitle}{\emph{Findings of the Association for Computational Linguistics: ACL 2023}}. \bibinfo{publisher}{Association for Computational Linguistics}, \bibinfo{address}{Toronto, Canada}, \bibinfo{pages}{1049--1065}.
\newblock
\urldef\tempurl%
\url{https://doi.org/10.18653/v1/2023.findings-acl.67}
\showDOI{\tempurl}


\bibitem[\protect\citeauthoryear{Huang, Yu, Ma, Zhong, Feng, et~al\mbox{.}}{Huang et~al\mbox{.}}{2023}]%
        {huang2023Hallucinationsurvey}
\bibfield{author}{\bibinfo{person}{Lei Huang}, \bibinfo{person}{Weijiang Yu}, \bibinfo{person}{Weitao Ma}, \bibinfo{person}{Weihong Zhong}, \bibinfo{person}{Zhangyin Feng}, {et~al\mbox{.}}} \bibinfo{year}{2023}\natexlab{}.
\newblock \bibinfo{title}{A Survey on Hallucination in Large Language Models: Principles, Taxonomy, Challenges, and Open Questions}.
\newblock
\newblock
\showeprint[arxiv]{2311.05232}~[cs.CL]


\bibitem[\protect\citeauthoryear{James}{James}{1890}]%
        {james1890principles}
\bibfield{author}{\bibinfo{person}{William James}.} \bibinfo{year}{1890}\natexlab{}.
\newblock \bibinfo{booktitle}{\emph{The Principles of Psychology}}.
\newblock \bibinfo{publisher}{Henry Holt and Company}.
\newblock


\bibitem[\protect\citeauthoryear{Jung, Qin, Welleck, Brahman, Bhagavatula, Bras, and Choi}{Jung et~al\mbox{.}}{2022}]%
        {jung2022maieutic}
\bibfield{author}{\bibinfo{person}{Jaehun Jung}, \bibinfo{person}{Lianhui Qin}, \bibinfo{person}{Sean Welleck}, \bibinfo{person}{Faeze Brahman}, \bibinfo{person}{Chandra Bhagavatula}, \bibinfo{person}{Ronan~Le Bras}, {and} \bibinfo{person}{Yejin Choi}.} \bibinfo{year}{2022}\natexlab{}.
\newblock \bibinfo{title}{Maieutic Prompting: Logically Consistent Reasoning with Recursive Explanations}.
\newblock
\newblock
\showeprint[arxiv]{2205.11822}~[cs.CL]


\bibitem[\protect\citeauthoryear{Kajić, Aygün, and Precup}{Kajić et~al\mbox{.}}{2020}]%
        {kajić2020learning}
\bibfield{author}{\bibinfo{person}{Ivana Kajić}, \bibinfo{person}{Eser Aygün}, {and} \bibinfo{person}{Doina Precup}.} \bibinfo{year}{2020}\natexlab{}.
\newblock \bibinfo{title}{Learning to cooperate: Emergent communication in multi-agent navigation}.
\newblock
\newblock
\showeprint[arxiv]{2004.01097}~[cs.LG]


\bibitem[\protect\citeauthoryear{Lange}{Lange}{1885}]%
        {lange1885mechanism}
\bibfield{author}{\bibinfo{person}{Carl Lange}.} \bibinfo{year}{1885}\natexlab{}.
\newblock \showarticletitle{The Mechanism of the Emotions}.
\newblock \bibinfo{journal}{\emph{The Classical Psychologists}} (\bibinfo{year}{1885}), \bibinfo{pages}{672--684}.
\newblock


\bibitem[\protect\citeauthoryear{Liu, Yuan, Fu, Jiang, Hayashi, and Neubig}{Liu et~al\mbox{.}}{2023}]%
        {PromptSurvey2023}
\bibfield{author}{\bibinfo{person}{Pengfei Liu}, \bibinfo{person}{Weizhe Yuan}, \bibinfo{person}{Jinlan Fu}, \bibinfo{person}{Zhengbao Jiang}, \bibinfo{person}{Hiroaki Hayashi}, {and} \bibinfo{person}{Graham Neubig}.} \bibinfo{year}{2023}\natexlab{}.
\newblock \showarticletitle{Pre-Train, Prompt, and Predict: A Systematic Survey of Prompting Methods in Natural Language Processing}.
\newblock \bibinfo{journal}{\emph{ACM Comput. Surv.}} \bibinfo{volume}{55}, \bibinfo{number}{9}, Article \bibinfo{articleno}{195} (\bibinfo{date}{jan} \bibinfo{year}{2023}), \bibinfo{numpages}{35}~pages.
\newblock


\bibitem[\protect\citeauthoryear{Manyika and Hsiao}{Manyika and Hsiao}{2023}]%
        {Bard2023}
\bibfield{author}{\bibinfo{person}{James Manyika} {and} \bibinfo{person}{Sissie Hsiao}.} \bibinfo{year}{2023}\natexlab{}.
\newblock \bibinfo{title}{{An overview of Bard: an early experiment with generative AI}}.
\newblock
\newblock
\urldef\tempurl%
\url{https://ai.google/static/documents/google-about-bard.pdf}
\showURL{%
\tempurl}


\bibitem[\protect\citeauthoryear{McHugh and Way}{McHugh and Way}{2018}]%
        {mchugh2018reasoning}
\bibfield{author}{\bibinfo{person}{Conor McHugh} {and} \bibinfo{person}{Jonathan Way}.} \bibinfo{year}{2018}\natexlab{}.
\newblock \showarticletitle{What is reasoning?}
\newblock \bibinfo{journal}{\emph{Mind}} \bibinfo{volume}{127}, \bibinfo{number}{505} (\bibinfo{year}{2018}), \bibinfo{pages}{167--196}.
\newblock


\bibitem[\protect\citeauthoryear{Newman-Toker, Nassery, and et~al.}{Newman-Toker et~al\mbox{.}}{2023}]%
        {Newman-Tokerbmjqs-2021-014130}
\bibfield{author}{\bibinfo{person}{David~E Newman-Toker}, \bibinfo{person}{Najlla Nassery}, {and} \bibinfo{person}{et al.}} \bibinfo{year}{2023}\natexlab{}.
\newblock \showarticletitle{Burden of serious harms from diagnostic error in the USA}.
\newblock \bibinfo{journal}{\emph{BMJ Quality \& Safety}} (\bibinfo{year}{2023}).
\newblock
\showISSN{2044-5415}
\showeprint{https://qualitysafety.bmj.com/content/early/2023/08/07/bmjqs-2021-014130.full.pdf}
\urldef\tempurl%
\url{https://qualitysafety.bmj.com/content/early/2023/08/07/bmjqs-2021-014130}
\showURL{%
\tempurl}


\bibitem[\protect\citeauthoryear{OpenAI}{OpenAI}{2021}]%
        {chatgpt}
\bibfield{author}{\bibinfo{person}{OpenAI}.} \bibinfo{year}{2021}\natexlab{}.
\newblock \bibinfo{title}{ChatGPT}.
\newblock
\newblock
\urldef\tempurl%
\url{https://openai.com/blog/chatgpt/}
\showURL{%
\tempurl}


\bibitem[\protect\citeauthoryear{OpenAI}{OpenAI}{2023a}]%
        {openai2023gpt4}
\bibfield{author}{\bibinfo{person}{OpenAI}.} \bibinfo{year}{2023}\natexlab{a}.
\newblock \bibinfo{title}{GPT-4 Technical Report}.
\newblock
\newblock
\showeprint[arxiv]{2303.08774}~[cs.CL]


\bibitem[\protect\citeauthoryear{OpenAI}{OpenAI}{2023b}]%
        {textDavinci003}
\bibfield{author}{\bibinfo{person}{OpenAI}.} \bibinfo{year}{2023}\natexlab{b}.
\newblock \showarticletitle{How do davinci and text-davinci-003 differ?}
\newblock \bibinfo{journal}{\emph{OpenAI Help Page}} (\bibinfo{year}{2023}).
\newblock
\urldef\tempurl%
\url{https://help.openai.com/en/articles/6643408-how-do-davinci-and-text-davinci-003-differ}
\showURL{%
\tempurl}


\bibitem[\protect\citeauthoryear{Patil}{Patil}{2020}]%
        {KaggleDSDD2020}
\bibfield{author}{\bibinfo{person}{Pranay Patil}.} \bibinfo{year}{2020}\natexlab{}.
\newblock \bibinfo{title}{{Kaggle Disease Symptoms Description Dataset}}.
\newblock
\newblock
\urldef\tempurl%
\url{https://www.kaggle.com/datasets/itachi9604/disease-symptom-description-dataset}
\showURL{%
\tempurl}


\bibitem[\protect\citeauthoryear{Paul and Elder}{Paul and Elder}{2007}]%
        {Paul2007CriticalTT}
\bibfield{author}{\bibinfo{person}{Richard~W. Paul} {and} \bibinfo{person}{Linda Elder}.} \bibinfo{year}{2007}\natexlab{}.
\newblock \showarticletitle{Critical Thinking: The Art of Socratic Questioning, Part III.}
\newblock \bibinfo{journal}{\emph{Journal of Developmental Education}}  \bibinfo{volume}{31} (\bibinfo{year}{2007}), \bibinfo{pages}{34--35}.
\newblock


\bibitem[\protect\citeauthoryear{Pearl}{Pearl}{1988}]%
        {pearl1988probabilistic}
\bibfield{author}{\bibinfo{person}{Judea Pearl}.} \bibinfo{year}{1988}\natexlab{}.
\newblock \bibinfo{booktitle}{\emph{Probabilistic Reasoning in Intelligent Systems: Networks of Plausible Inference}}.
\newblock \bibinfo{publisher}{Morgan Kaufmann}.
\newblock
\showISBNx{0-934613-73-7}


\bibitem[\protect\citeauthoryear{Pearl}{Pearl}{2009}]%
        {pearl2009causality}
\bibfield{author}{\bibinfo{person}{Judea Pearl}.} \bibinfo{year}{2009}\natexlab{}.
\newblock \bibinfo{booktitle}{\emph{Causality: Models, Reasoning and Inference} (\bibinfo{edition}{2nd} ed.)}.
\newblock \bibinfo{publisher}{Cambridge University Press}.
\newblock
\showISBNx{978-0521895606}


\bibitem[\protect\citeauthoryear{Sap, LeBras, Fried, and Choi}{Sap et~al\mbox{.}}{2023}]%
        {sap2023neural}
\bibfield{author}{\bibinfo{person}{Maarten Sap}, \bibinfo{person}{Ronan LeBras}, \bibinfo{person}{Daniel Fried}, {and} \bibinfo{person}{Yejin Choi}.} \bibinfo{year}{2023}\natexlab{}.
\newblock \bibinfo{title}{Neural Theory-of-Mind? On the Limits of Social Intelligence in Large LMs}.
\newblock
\newblock
\showeprint[arxiv]{2210.13312}~[cs.CL]


\bibitem[\protect\citeauthoryear{Schick and Sch{\"u}tze}{Schick and Sch{\"u}tze}{2020}]%
        {Schick2020ExploitingCF}
\bibfield{author}{\bibinfo{person}{Timo Schick} {and} \bibinfo{person}{Hinrich Sch{\"u}tze}.} \bibinfo{year}{2020}\natexlab{}.
\newblock \showarticletitle{Exploiting Cloze-Questions for Few-Shot Text Classification and Natural Language Inference}. In \bibinfo{booktitle}{\emph{Conference of the European Chapter of the Association for Computational Linguistics}}.
\newblock


\bibitem[\protect\citeauthoryear{Sclar, Kumar, West, Suhr, Choi, and Tsvetkov}{Sclar et~al\mbox{.}}{2023}]%
        {sclar2023minding}
\bibfield{author}{\bibinfo{person}{Melanie Sclar}, \bibinfo{person}{Sachin Kumar}, \bibinfo{person}{Peter West}, \bibinfo{person}{Alane Suhr}, \bibinfo{person}{Yejin Choi}, {and} \bibinfo{person}{Yulia Tsvetkov}.} \bibinfo{year}{2023}\natexlab{}.
\newblock \bibinfo{title}{Minding Language Models' (Lack of) Theory of Mind: A Plug-and-Play Multi-Character Belief Tracker}.
\newblock
\newblock
\showeprint[arxiv]{2306.00924}~[cs.CL]


\bibitem[\protect\citeauthoryear{Team, Anil, Borgeaud, Wu, Alayrac, Yu, Soricut, Schalkwyk, et~al\mbox{.}}{Team et~al\mbox{.}}{2023}]%
        {geminiteam2023gemini}
\bibfield{author}{\bibinfo{person}{Gemini Team}, \bibinfo{person}{Rohan Anil}, \bibinfo{person}{Sebastian Borgeaud}, \bibinfo{person}{Yonghui Wu}, \bibinfo{person}{Jean-Baptiste Alayrac}, \bibinfo{person}{Jiahui Yu}, \bibinfo{person}{Radu Soricut}, \bibinfo{person}{Johan Schalkwyk}, {et~al\mbox{.}}} \bibinfo{year}{2023}\natexlab{}.
\newblock \bibinfo{title}{Gemini: A Family of Highly Capable Multimodal Models}.
\newblock
\newblock
\showeprint[arxiv]{2312.11805}~[cs.CL]


\bibitem[\protect\citeauthoryear{Thoppilan, Freitas, Hall, and et~al.}{Thoppilan et~al\mbox{.}}{2022}]%
        {thoppilan2022lamda}
\bibfield{author}{\bibinfo{person}{Romal Thoppilan}, \bibinfo{person}{Daniel~De Freitas}, \bibinfo{person}{Jamie Hall}, {and} \bibinfo{person}{et al.}} \bibinfo{year}{2022}\natexlab{}.
\newblock \bibinfo{title}{LaMDA: Language Models for Dialog Applications}.
\newblock
\newblock
\showeprint[arxiv]{2201.08239}~[cs.CL]


\bibitem[\protect\citeauthoryear{Touvron, Martin, Stone, and et~al.}{Touvron et~al\mbox{.}}{2023}]%
        {touvron2023llama}
\bibfield{author}{\bibinfo{person}{Hugo Touvron}, \bibinfo{person}{Louis Martin}, \bibinfo{person}{Kevin Stone}, {and} \bibinfo{person}{et al.}} \bibinfo{year}{2023}\natexlab{}.
\newblock \bibinfo{title}{Llama 2: Open Foundation and Fine-Tuned Chat Models}.
\newblock
\newblock
\showeprint[arxiv]{2307.09288}~[cs.CL]


\bibitem[\protect\citeauthoryear{Tsao}{Tsao}{2023}]%
        {SocraPlan2023}
\bibfield{author}{\bibinfo{person}{Wen-Kwang Tsao}.} \bibinfo{year}{2023}\natexlab{}.
\newblock \showarticletitle{{Multi-Agent Reasoning with Large Language Models for Effective Corporate Planning}}. In \bibinfo{booktitle}{\emph{The $10^{th}$ International Conf. on Computational Science and Computational Intelligence}}.
\newblock


\bibitem[\protect\citeauthoryear{Valmeekam, Olmo, Sreedharan, and Kambhampati}{Valmeekam et~al\mbox{.}}{2022}]%
        {valmeekam2022large}
\bibfield{author}{\bibinfo{person}{Karthik Valmeekam}, \bibinfo{person}{Alberto Olmo}, \bibinfo{person}{Sarath Sreedharan}, {and} \bibinfo{person}{Subbarao Kambhampati}.} \bibinfo{year}{2022}\natexlab{}.
\newblock \showarticletitle{Large Language Models Still Can't Plan (A Benchmark for LLMs on Planning and Reasoning about Change)}. In \bibinfo{booktitle}{\emph{NeurIPS 2022 Foundation Models for Decision Making Workshop}}.
\newblock


\bibitem[\protect\citeauthoryear{Wason and Johnson-Laird}{Wason and Johnson-Laird}{1972}]%
        {wason1972psychology}
\bibfield{author}{\bibinfo{person}{Peter~Cathcart Wason} {and} \bibinfo{person}{Philip~Nicholas Johnson-Laird}.} \bibinfo{year}{1972}\natexlab{}.
\newblock \bibinfo{booktitle}{\emph{Psychology of reasoning: Structure and content}}. Vol.~\bibinfo{volume}{86}.
\newblock \bibinfo{publisher}{Harvard University Press}.
\newblock


\bibitem[\protect\citeauthoryear{Wei, Wang, Schuurmans, Bosma, Ichter, Xia, Chi, Le, and Zhou}{Wei et~al\mbox{.}}{2022}]%
        {ChainExtended2023}
\bibfield{author}{\bibinfo{person}{Jason Wei}, \bibinfo{person}{Xuezhi Wang}, \bibinfo{person}{Dale Schuurmans}, \bibinfo{person}{Maarten Bosma}, \bibinfo{person}{Brian Ichter}, \bibinfo{person}{Fei Xia}, \bibinfo{person}{Ed Chi}, \bibinfo{person}{Quoc Le}, {and} \bibinfo{person}{Denny Zhou}.} \bibinfo{year}{2022}\natexlab{}.
\newblock \bibinfo{title}{Chain-of-Thought Prompting Elicits Reasoning in Large Language Models}.
\newblock
\newblock
\urldef\tempurl%
\url{https://doi.org/10.48550/ARXIV.2201.11903}
\showDOI{\tempurl}


\bibitem[\protect\citeauthoryear{Wikipedia}{Wikipedia}{2023}]%
        {SocraticMethidWiki}
\bibfield{author}{\bibinfo{person}{Wikipedia}.} \bibinfo{year}{2023}\natexlab{}.
\newblock \bibinfo{title}{Socratic method}.
\newblock
\newblock
\urldef\tempurl%
\url{https://en.wikipedia.org/wiki/Socratic_method}
\showURL{%
\tempurl}


\bibitem[\protect\citeauthoryear{Wrenn}{Wrenn}{2023}]%
        {PaltoRepublicURL}
\bibfield{author}{\bibinfo{person}{Chase~B. Wrenn}.} \bibinfo{year}{2023}\natexlab{}.
\newblock \bibinfo{title}{Internet Encyclopedia of Philosophy}.
\newblock
\newblock
\urldef\tempurl%
\url{https://iep.utm.edu/republic/}
\showURL{%
\tempurl}


\bibitem[\protect\citeauthoryear{Yao, Yu, Zhao, Shafran, Griffiths, Cao, and Narasimhan}{Yao et~al\mbox{.}}{2023}]%
        {yao2023tree}
\bibfield{author}{\bibinfo{person}{Shunyu Yao}, \bibinfo{person}{Dian Yu}, \bibinfo{person}{Jeffrey Zhao}, \bibinfo{person}{Izhak Shafran}, \bibinfo{person}{Thomas~L. Griffiths}, \bibinfo{person}{Yuan Cao}, {and} \bibinfo{person}{Karthik Narasimhan}.} \bibinfo{year}{2023}\natexlab{}.
\newblock \bibinfo{title}{Tree of Thoughts: Deliberate Problem Solving with Large Language Models}.
\newblock
\newblock
\showeprint[arxiv]{2305.10601}~[cs.CL]


\bibitem[\protect\citeauthoryear{Yuan, Yuan, Li, Dong, Lu, Tan, Zhou, and Zhou}{Yuan et~al\mbox{.}}{2023}]%
        {yuan2023scaling}
\bibfield{author}{\bibinfo{person}{Zheng Yuan}, \bibinfo{person}{Hongyi Yuan}, \bibinfo{person}{Chengpeng Li}, \bibinfo{person}{Guanting Dong}, \bibinfo{person}{Keming Lu}, \bibinfo{person}{Chuanqi Tan}, \bibinfo{person}{Chang Zhou}, {and} \bibinfo{person}{Jingren Zhou}.} \bibinfo{year}{2023}\natexlab{}.
\newblock \bibinfo{title}{Scaling Relationship on Learning Mathematical Reasoning with Large Language Models}.
\newblock
\newblock
\showeprint[arxiv]{2308.01825}~[cs.CL]


\bibitem[\protect\citeauthoryear{Zeng}{Zeng}{2022}]%
        {SocraticModels-Google2022}
\bibfield{author}{\bibinfo{person}{Andy et~al Zeng}.} \bibinfo{year}{2022}\natexlab{}.
\newblock \bibinfo{title}{Socratic Models: Composing Zero-Shot Multimodal Reasoning with Language}.
\newblock
\newblock


\bibitem[\protect\citeauthoryear{Zhang, Li, Meng, Chang, and den Broeck}{Zhang et~al\mbox{.}}{2022}]%
        {zhang2022paradox}
\bibfield{author}{\bibinfo{person}{Honghua Zhang}, \bibinfo{person}{Liunian~Harold Li}, \bibinfo{person}{Tao Meng}, \bibinfo{person}{Kai-Wei Chang}, {and} \bibinfo{person}{Guy~Van den Broeck}.} \bibinfo{year}{2022}\natexlab{}.
\newblock \bibinfo{title}{On the Paradox of Learning to Reason from Data}.
\newblock
\newblock
\showeprint[arxiv]{2205.11502}~[cs.CL]


\bibitem[\protect\citeauthoryear{Zhang, Yang, Yuan, and Yao}{Zhang et~al\mbox{.}}{2023}]%
        {zhang2023cumulative}
\bibfield{author}{\bibinfo{person}{Yifan Zhang}, \bibinfo{person}{Jingqin Yang}, \bibinfo{person}{Yang Yuan}, {and} \bibinfo{person}{Andrew Chi-Chih Yao}.} \bibinfo{year}{2023}\natexlab{}.
\newblock \bibinfo{title}{Cumulative Reasoning with Large Language Models}.
\newblock
\newblock
\showeprint[arxiv]{2308.04371}~[cs.AI]


\end{thebibliography}
